\pdfoutput=1

\documentclass[11pt]{article}

\usepackage[final]{acl}
\usepackage{verbatim}
\usepackage{times}
\usepackage{latexsym}

\usepackage[T1]{fontenc}

\usepackage[utf8]{inputenc}

\usepackage{microtype}

\usepackage{inconsolata}

\usepackage{graphicx}
\usepackage{cuted} 
\usepackage{booktabs}   
\usepackage{multirow}   
\usepackage{adjustbox} 
\usepackage{placeins}
\usepackage{tabularx}
\usepackage{makecell}
\usepackage{amsmath}
\usepackage{algorithm2e}
\usepackage{float}
\usepackage{placeins}



\title{NormGenesis: Multicultural Dialogue Generation via Exemplar-Guided Social Norm Modeling and Violation Recovery}

\author{Minki Hong$^{1}$ \and Jangho Choi$^{1}$ \and Jihie Kim$^{1}$\thanks{~Corresponding author: \texttt{jihie.kim@dgu.edu}}\\[3pt]
$^{1}$Department of Computer Science and Artificial Intelligence, Dongguk University \\
    \texttt{\{jackyh1, 2025120382\}@dgu.ac.kr, jihie.kim@dgu.edu}
}

\begin{document}
\maketitle
\begin{abstract}
Social norms govern culturally appropriate behavior in communication, enabling dialogue systems to produce responses that are not only coherent but also socially acceptable. We present NormGenesis, a multicultural framework for generating and annotating socially grounded dialogues across English, Chinese, and Korean. To model the dynamics of social interaction beyond static norm classification, we propose a novel dialogue type, Violation-to-Resolution (V2R), which models the progression of conversations following norm violations through recognition and socially appropriate repair. To improve pragmatic consistency in underrepresented languages, we implement an exemplar-based iterative refinement early in the dialogue synthesis process. This design introduces alignment with linguistic, emotional, and sociocultural expectations before full dialogue generation begins. Using this framework, we construct a dataset of 10,800 multi-turn dialogues annotated at the turn level for norm adherence, speaker intent, and emotional response. Human and LLM-based evaluations demonstrate that NormGenesis significantly outperforms existing datasets in refinement quality, dialogue naturalness, and generalization performance. We show that models trained on our V2R-augmented data exhibit improved pragmatic competence in ethically sensitive contexts. Our work establishes a new benchmark for culturally adaptive dialogue modeling and provides a scalable methodology for norm-aware generation across linguistically and culturally diverse languages.
\end{abstract}

\section{Introduction}
\label{Sec:1}
\begin{figure}[t]
\centering
\includegraphics[height=10.7cm]{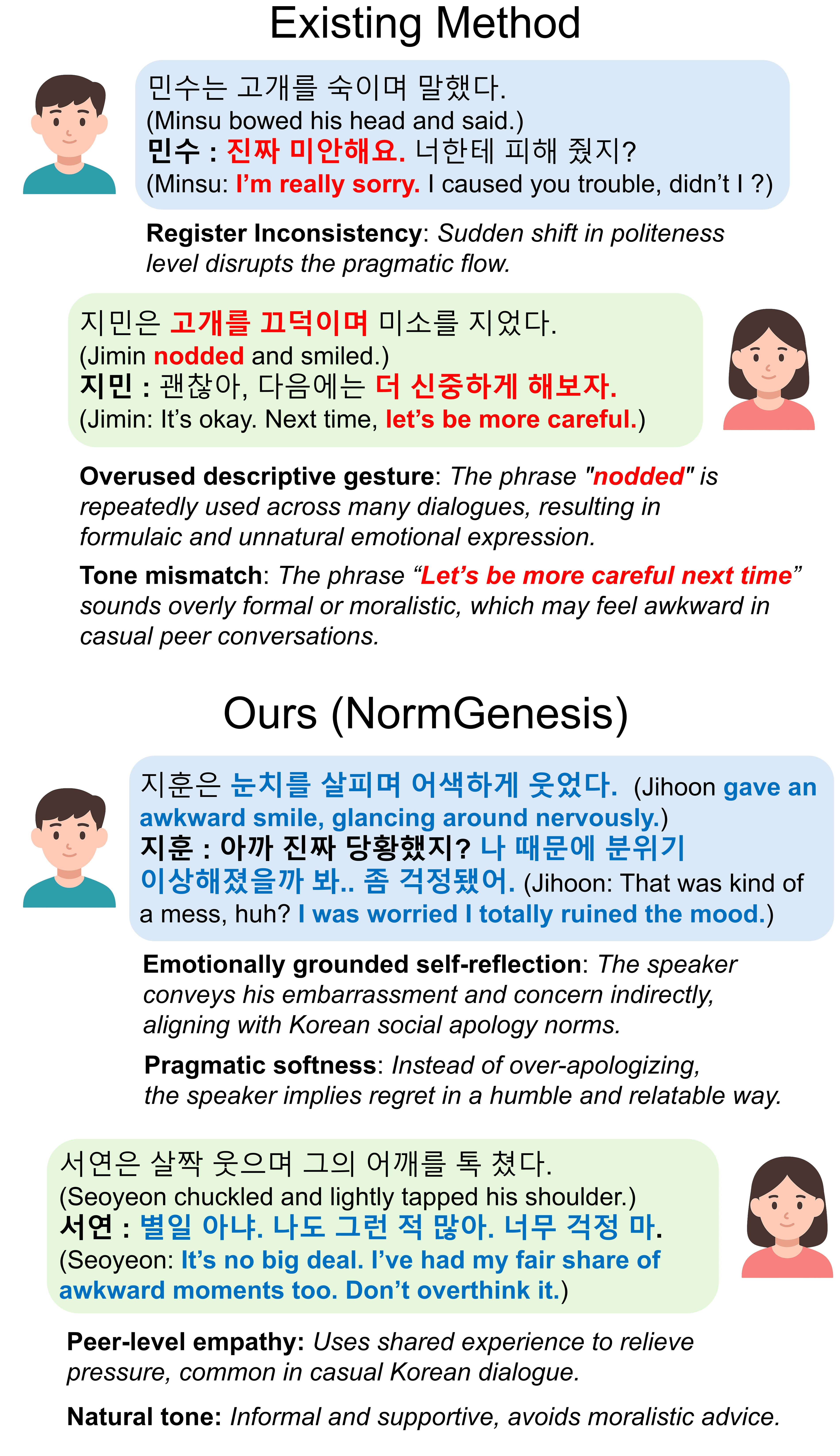}
\caption{Comparison of generation outputs in Korean. Prior methods~\cite{li2023normdial} produce pragmatically inconsistent responses, including honorific misuse and unnatural tone (highlighted in red). In contrast, our framework yields culturally and pragmatically coherent outputs (highlighted in blue).}
\label{fig:problem}
\end{figure} 

Social norms are culturally defined expectations that guide appropriate behavior in specific contexts~\cite{elster2006fairness, malle2014theory}. In human communication, social norms support politeness, empathy, and social harmony. For dialogue systems, aligning with social norms enables responses that transcend syntactic correctness or task completion, contributing to pragmatic and interpersonal appropriateness~\cite{kim2022prosocialdialog}. As conversational agents are increasingly deployed in socially embedded and open-domain settings, the ability to recognize and adhere to cultural norms has become a critical indicator of social and pragmatic competence~\cite{zhan2023socialdial}.

Recent studies have integrated social norms into dialogue datasets and language models. Prior works have explored moral reasoning in language models~\cite{forbes2020social}, norm-based labeling~\cite{li2023normdial}, emotion-informed norm interpretation~\cite{zhan-etal-2024-renovi}, and cross-cultural generalization~\cite{rao-etal-2025-normad}. While these efforts lay foundational groundwork for norm-aware generation, they primarily focus on English, a high-resource language. Although Chinese has received growing attention, the modeling of culturally appropriate behavior remains underdeveloped for low-resource languages. This limitation is especially pronounced in Korean, where existing models frequently exhibit inconsistencies in honorific usage, inadequate emotional alignment, and misrepresentation of role-based social dynamics~\cite{jang-etal-2024-kodialogbench, lee-etal-2024-kornat}, as illustrated in Figure~\ref{fig:problem}.

To address the cultural and pragmatic limitations of existing dialogue datasets, we present a multicultural framework for generating and refining socially grounded dialogues across English, Chinese, and Korean. While English benefits from extensive data and modeling maturity, low-resource languages generation remains challenged by pragmatic mismatches, especially in tone and formality~\cite{zhong2024opportunitieschallengeslargelanguage}. We mitigate this gap through an exemplar-based iterative refinement strategy. Given a target scenario, the system retrieves semantically and structurally aligned exemplars, using features such as intent, emotional tone, and discourse patterns (e.g., speaker roles and adjacency). These exemplars guide revision to ensure cultural alignment without requiring large-scale human annotation.
We further introduce a novel dialogue category, \textit{Violation-to-Resolution} (V2R), which captures how speakers recover from norm violations through contextually appropriate repair. This enables the modeling of pragmatically dynamic interactions that reflect both norm compliance and social recovery mechanisms. Leveraging our framework, we construct a dataset of 10,800 high-quality dialogues annotated at the turn level with norm adherence, violation, speaker intent, and emotional response, grounded in dialogue act theory~\cite{bunt-etal-2020-iso}. 
We evaluate our approach through both human and LLM-based assessments of refinement quality, dialogue fluency, social appropriateness, and generalization. Experimental results show that models trained on our data significantly outperform existing baselines in socially complex and emotionally sensitive scenarios. These findings demonstrate the efficacy of our framework for enabling culturally adaptive dialogue generation across typologically diverse languages.

Our main contributions are as follows:
\begin{enumerate}
    \item We present a multicultural framework for generating socially grounded dialogues in English, Chinese, and Korean. To address cultural and pragmatic degradation in low-resource settings, we propose an exemplar-based iterative refinement strategy using semantically relevant exemplars.

\item We propose a novel dialogue type, \textit{Violation-to-Resolution} (V2R), that models how norm violations are followed by socially appropriate repair, enabling the representation of dynamic and culturally meaningful interaction patterns.

\item We construct a dataset of 10,800 multicultural dialogues with turn-level annotations for norm adherence, speaker intent, and emotional response, grounded in dialogue act theory.

\item We show that models trained on our dataset outperform prior resources in norm alignment, emotional coherence, and repair, as confirmed by both human and automatic evaluations.
\end{enumerate}

\section{Related Work}
\label{Sec:2}
\subsection{Social Norms in Dialogue Systems}
\label{Sec:2.1}
Integrating social norms into dialogue systems is critical for generating contextually appropriate and socially coherent responses. Early work~\cite{forbes2020social} provided normative signals via moral judgments but lacked dialogue structure. Later efforts~\cite{li2023normdial} added norm annotations to multi-turn dialogues, focusing on adherence classification without modeling responses to violations. Recent work~\cite{zhan-etal-2024-renovi} has begun modeling norm repair in dialogue. However, it remains monolingual and lacks fine-grained annotations capturing speaker intent and emotional response. 

To bridge this gap, we introduce \textit{Violation-to-Resolution} as a distinct response type, capturing repair strategies such as apology and explanation. Also, our framework supports this with turn-level annotations of communicative intent and emotional state, enabling more nuanced and socially competent generation.

\begin{figure*}[t]
\centering
\includegraphics[width=\textwidth]{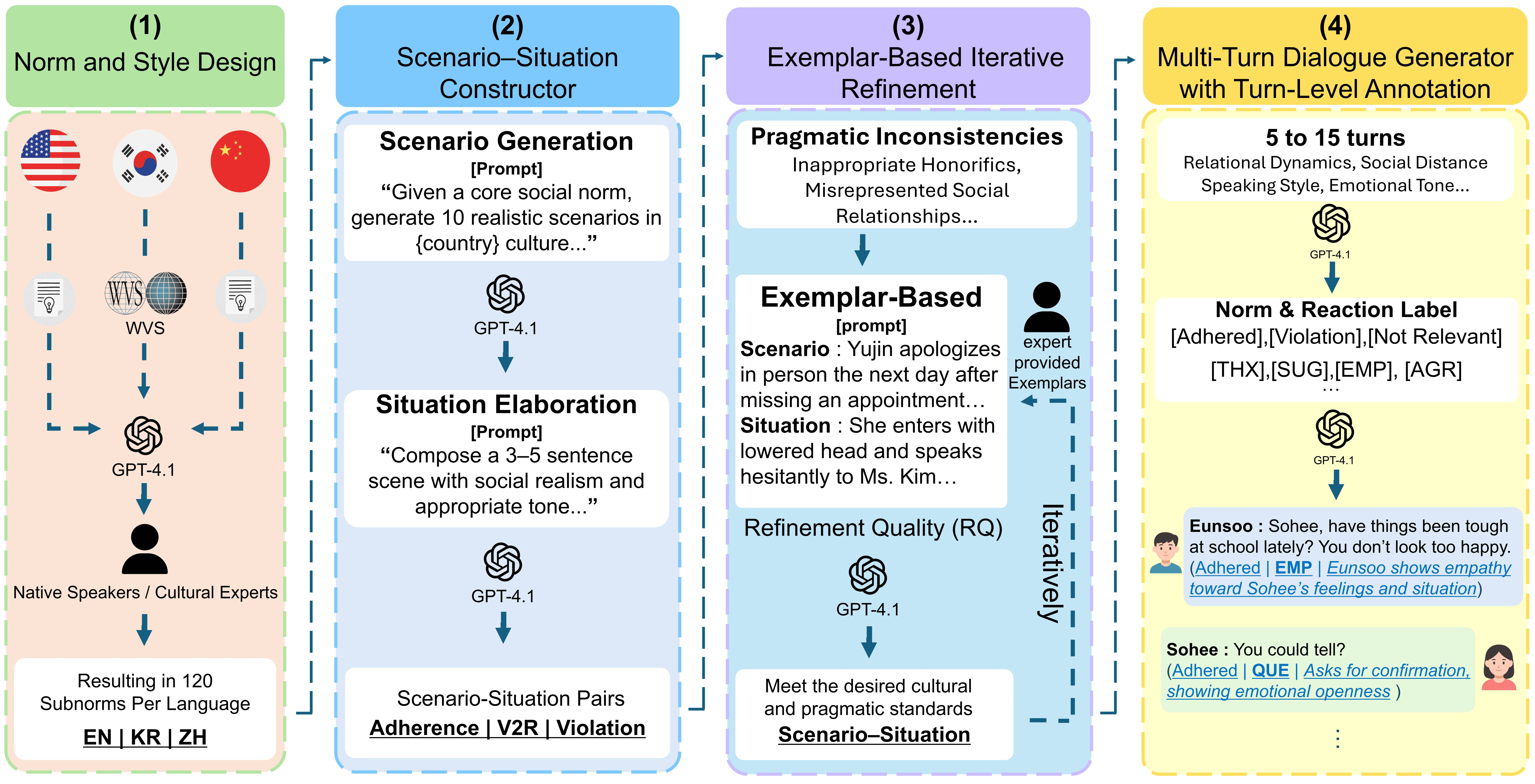}
\caption{\textbf{NormGenesis Overview}. Our framework consists of four stages: (1) culturally grounded norm and style design, (2) scenario–situation construction across norm adherence, violation, and resolution types, (3) exemplar-based iterative refinement using semantically aligned exemplars, and (4) multi-turn dialogue generation with turn-level annotations. Each stage is evaluated and refined iteratively to ensure pragmatic consistency and cultural alignment, as described in Section~\ref{Sec:3}. The RQ evaluation protocol is described in Section~\ref{Sec:4.2}.}
\label{fig:2}
\end{figure*}

\subsection{Prompt-Based and Exemplar-Guided Generation}
\label{Sec:2.2}
The emergence of LLMs has enabled prompt-based synthetic data generation through techniques such as in-context learning and prompt tuning. While effective in high-resource settings, these methods often fail to capture cultural and emotional nuance in low-resource languages~\cite{cahyawijaya-etal-2024-llms, pranida2025syntheticdatagenerationculturally}. Recent approaches~\cite{nguyen2024democratizing, ghosal-etal-2025-promptrefine} explore exemplar-based generation, but typically rely on fixed examples and prioritizes fluency over pragmatic fit. SADAS~\cite{hua-etal-2024-assistive} also investigated norm interventions through exemplar-based remediation. However, it applies these exemplars reactively after violations occur in negotiation dialogues, remaining limited to post-hoc repair.

To address these limitations, we propose an exemplar-based iterative refinement strategy. Unlike few-shot prompting, which uses static exemplars at inference time, our method dynamically selects semantically and structurally relevant examples to guide generation. This process improves linguistic coherence and cultural alignment while enabling data bootstrapping without large-scale human annotation, particularly in low-resource languages such as Korean.

\subsection{Culturally Adaptive Dialogue}
\label{Sec:2.3}
As conversational AI systems expand to diverse cultural contexts, cultural adaptability becomes essential. Early multilingual datasets, such as \textsc{BiToD}\cite{lin2021bitodbilingualmultidomaindataset} and \textsc{Multi3WOZ}\cite{hu2023multi3wozmultilingualmultidomainmultiparallel}, introduced bilingual dialogues but lacked cultural annotations and failed to capture pragmatic nuance due to reliance on translation. Recent works, including \textsc{CARE}\cite{guo2025carealigninglanguagemodels} and \textsc{CulturePark}\cite{li2024cultureparkboostingcrossculturalunderstanding}, incorporate cultural preferences and simulates cross-cultural interactions with LLMs. However, these resources focus on high-resource languages and lack the fine-grained, turn-level annotations needed for culturally appropriate generation.

This gap is pronounced in underrepresented languages like Korean, where honorifics and relational pragmatics are central. To address this, we propose a multicultural framework with language-specific subnorms and turn-level annotations. Using exemplar-based iterative refinement, our method enhances cultural and pragmatic consistency, particularly in low-resource settings.

\section{Method}
\label{Sec:3}
We introduce NormGenesis, a multicultural framework for generating and refining socially grounded dialogues in American English, Chinese, and Korean. While we broadly refer to "English" in our framework, it is important to clarify that the dataset primarily reflects U.S.-based social norms, derived from American corpora and sociocultural frameworks. This specification ensures cultural precision and avoids conflating diverse normative practices across other English-speaking contexts (e.g., British, Australian, Canadian). The framework consists of four core stages, each responsible for modeling social communication: (1) norm and style design, (2) scenario-situation constructor, (3) exemplar-based iterative refinement engine, and (4) multi-turn dialogue generator with turn-level annotation. Figure~\ref{fig:2} illustrates the overall pipeline and stage flow. Also, the complete algorithmic workflow of our framework is provided in Appendix~\ref{appendix:c.1}.

\begin{table}[t]
\small
\centering
\begin{tabular}{ll}
\toprule
Apology & Compliment \\
\midrule
Condolence & Criticism \\
\midrule
Empathy & Greeting \\
\midrule
Leave-taking & Persuasion \\
\midrule
Request & Respect \\
\midrule
Responding to Compliments & Thanks\\
\bottomrule
\end{tabular}
\caption{Social norm categories used in this study.}
\label{tab:norm-category}
\end{table}

\subsection{Norm and Style Design}
\label{Sec:3.1}
We construct a taxonomy of 12 conversational social norm categories by extending the 10 categories proposed in~\cite{li2023normdial} with two additional types: \textit{Empathy} and \textit{Respect}, motivated by prior work on dialogic functions~\cite{stolcke2000dialogue, bunt-etal-2020-iso}. The complete set is listed in Table~\ref{tab:norm-category}. For each category, we define 10 culturally grounded subnorms per target language. To generate Korean-specific subnorms, we prompt an LLM with value-centric responses from the World Values Survey (WVS Wave 7, South Korea)~\cite{wvs2022round7}. English and Chinese subnorms are adapted from~\cite{li2023normdial} through LLM-guided alignment with the Korean outputs. All subnorms are validated by native speakers or cultural experts to ensure fluency and cultural plausibility, yielding 120 subnorms per language (Figure~\ref{fig:2} (1)). We also define pragmatic and stylistic parameters that guide both scenario construction and dialogue generation. These include tone(formal vs. casual), honorific usage, relational distance (peer vs. hierarchical), and emotional alignment. Representative uses are shown in Table~\ref{tab:refinement}, with prompt templates and specifications provided in Appendix~\ref{appendix:c}.

\subsection{Scenario-Situation Constructor}
\label{Sec:3.2}
For each subnorm, we construct a scenario–situation pair consisting of: (a) a scenario that provides a concise, real-world context, and (b) a situation that expands upon the scenario with 3–5 sentences specifying relational roles, emotional states, and stylistic features such as tone and honorifics. Each instance is labeled as one of three interaction types: Norm Adherence, Norm Violation, or Violation-to-Resolution (V2R). While the first two reflect conventional norm conformity or transgression, V2R models post-violation repair strategies, capturing core aspects of interactional competence~\cite{goffman2017interaction, feine2019taxonomy}. Despite its importance, V2R remains largely absent from prior norm-based dialogue datasets. To our knowledge, this is the first work to formally define and incorporate V2R into social dialogue modeling. Examples of each type are shown in Appendix~\ref{appendix:d}. As shown in Figure~\ref{fig:2} (2), scenarios are first generated by prompting an LLM with a subnorm and interaction type. These are then expanded into situations via a second prompt enriched with interpersonal and emotional cues. Notably, while English outputs are generally fluent, Korean and Chinese generations often contain pragmatic inconsistencies (e.g., tone mismatch, incorrect honorifics). These issues motivate the exemplar-based refinement described in Section~\ref{Sec:3.3}.

\subsection{Exemplar-Based Iterative Refinement}
\label{Sec:3.3}
Unlike prior approaches that rely on post-hoc filtering or manual correction after dialogue generation~\cite{lambert2024self, occhipinti-etal-2024-fine}, our framework introduces an upstream refinement mechanism at the scenario–situation level, enabling early enforcement of cultural and pragmatic constraints. For each norm category, we manually curate a small set of high-quality exemplars that reflect culturally grounded and stylistically appropriate behaviors. Rather than using static prompts, the model retrieves semantically and structurally similar exemplars based on communicative intent, emotional tone, and discourse patterns (e.g., speaker roles and adjacency), guiding the revision process without large-scale human annotation.

To determine whether further refinement is necessary, we implement an iterative loop using our \textit{Refinement Quality} (RQ) protocol (Section~\ref{Sec:4.2}), which evaluates the quality of the revised output in comparison to the original input. The model receives a triplet:
\begin{equation}
\left( \text{Input}_{\text{orig}},\ \text{Output}_{\text{refined}},\ \text{Score}_{\text{qual}} \right)
\end{equation}
enabling it to assess the social and stylistic adequacy of the revision and decide whether to continue refinement. Here, $\text{Input}_{\text{orig}}$ denotes the original scenario–situation pair, $\text{Output}_{\text{refined}}$ is the revised output from exemplar-based prompting, and $\text{Score}_{\text{qual}}$ is a scalar computed via the RQ protocol.

This early integration mitigates quality bottlenecks common in generation-first pipelines and ensures consistent sociocultural alignment prior to dialogue construction. The effectiveness of our approach is illustrated in Appendix~\ref{appendix:d} through representative before-and-after examples across languages and norm categories, with full evaluation results reported in Section~\ref{Sec:5.1}.

\subsection{Multi-Turn Dialogue Generator with Turn-Level Annotation}
\label{Sec:3.4}
After refinement, each scenario–situation pair is expanded into a multi-turn dialogue (5–15 turns), resulting in socially and contextually appropriate interactions. Each utterance is annotated with (a) norm adherence, (b) speaker reaction including intent and emotional state, and (c) justification for the assigned label. This structure enables fine-grained modeling of social dynamics by identifying norm compliance and explaining speaker behavior. Reaction labels, grounded in dialogue act theory~\cite{stolcke2000dialogue}, are assigned via LLM-based prompting and expert verification, as detailed in Appendix~\ref{appendix:c.7}. The resulting dataset supports norm reasoning and socially sensitive dialogue modeling across cultures. Representative examples of the generated dialogues and annotations are provided in Appendix~\ref{appendix:d}.

\section{Evaluation Framework}
\label{Sec:4}
\subsection{Datasets and Experimental Conditions}
\label{Sec:4.1}
\paragraph{Dataset Composition and Baselines.}
We construct a multicultural dataset in English, Chinese, and Korean, with 1,200 instances per language across three interaction types: Norm Adherence, Norm Violation, and \textit{Violation-to-Resolution} (V2R), resulting in 10,800 instances overall. We conducted our experiments using four NVIDIA A100 GPUs over a duration of eight hours.

For the baselines, we compare against the following existing resources:
\begin{itemize}
    \item \textsc{NormDial}~\cite{li2023normdial}: A bilingual English–Chinese corpus with turn-level norm adherence and violation annotations.
    \item \textsc{SODA}~\cite{kim2023soda}: An English corpus of socially appropriate dialogues grounded in social commonsense.
\end{itemize}

\paragraph{Model Configuration and Evaluation Protocol.}
All scenario–situation refinements and dialogue generations are conducted using GPT-4.1~\cite{openai2025models}, guided by a small set of expert-curated exemplars derived from manually revised model outputs. These exemplars reflect culturally and pragmatically appropriate responses and are used to steer iterative refinement, particularly in low-resource languages. Illustrative examples and refinement prompts are provided in Appendix~\ref{appendix:c.5}. Automatic evaluations are performed using GPT-4o in a zero-shot setting with metric-specific prompts as described in Appendix~\ref{appendix:e}.

For downstream experiments (Section~\ref{Sec:5}), we fine-tune both closed and open-source language models: GPT-4o-mini as a closed model, and LLaMA-3-8B~\cite{llama3modelcard}, Qwen-2.5-14B, and Qwen-2.5-32B~\cite{qwen2.5} as open-source baselines. All models are trained under identical configurations for each language to ensure comparability. Human evaluations were conducted by native Chinese and Korean speakers who were selected for their linguistic fluency and cultural familiarity. To assess model performance in low-resource cultural settings, we recruited six independent graduate students (four Korean, two Chinese) as annotators. For each evaluation, we randomly sampled 100 dialogues per type, and annotators rated the outputs on fluency, relevance, and social norm adherence using a Likert-scale judgment. These human assessments were further complemented with automatic evaluations based on our proposed metrics. Detailed procedures are provided in Appendix~\ref{appendix:e}.

\subsection{Evaluation Objectives and Design}
\label{Sec:4.2}
We evaluate our framework along three axes:
\begin{enumerate}
    \item \textbf{Refinement Quality (RQ)}: Does our refinement method improve generation quality in low-resource languages?
    \item \textbf{Dialogue Quality(DQ)}: Do generated dialogues align with norms and pragmatic expectations?
    \item \textbf{Generalization Quality (GQ)}: Do our models outperform baselines in quality and human preference?
\end{enumerate}

Evaluation templates for both LLM and human assessments are detailed in Appendix~\ref{appendix:e}, along with detailed evaluation description and guidelines.

\begin{table}[t]
\small
\centering
\begin{tabularx}{\linewidth}{lX}
\toprule
\textbf{Criterion} & \textbf{Description} \\
\midrule
\textbf{Norm Alignment} & Adherence to the intended social norm \\
\textbf{Language Quality} & Grammaticality and fluency \\
\textbf{Semantic Fidelity} & Preservation of original intent \\
\bottomrule
\end{tabularx}
\caption{Evaluation criteria used for refinement quality assessment. Scoring uses a 5-point Likert scale with both LLM and expert annotators. Detailed prompt templates for each evaluation criterion are included in Appendix~\ref{appendix:e.1}}
\label{tab:refinement-criteria}
\end{table}

\paragraph{Refinement Quality (RQ)}
We compare scenario-situation pairs before and after refinement in Korean and Chinese. Table~\ref{tab:refinement-criteria} summarizes the three key dimensions used to evaluate refinement quality.

\begin{table}[t]
\small
\centering
\begin{tabularx}{\linewidth}{lX}
\toprule
\textbf{Criterion} & \textbf{Description} \\
\midrule
\makecell[l]{Consistency} & Logical flow across turns \\
\midrule
\makecell[l]{Naturalness} & Fluency and human-likeness \\
\midrule
\makecell[l]{Relevance} & Context-appropriate responses \\
\midrule
\makecell[l]{Emotion\\Appropriateness} & Tone aligned with context \\
\midrule
\makecell[l]{Social Norm\\Appropriateness} & Cultural and normative compliance \\
\midrule
\makecell[l]{Scenario\\Coherence} & Semantic alignment with the scenario \\
\bottomrule
\end{tabularx}
\caption{Evaluation criteria used for dialogue quality (DQ). Detailed are included in Appendix~\ref{appendix:e.2}}
\label{tab:dq-criteria}
\end{table}

\begin{table*}[t]
\centering
\small
\begin{adjustbox}{max width=\linewidth}
\begin{tabular}{llccc}
\toprule
\textbf{Language} & \textbf{Condition} & \textbf{Norm Align.} & \textbf{Linguistic Quality} & \textbf{Semantic Fidelity} \\
\midrule
\multirow{2}{*}{Korean} 
& Initial  & 4.577 & 3.589 & N/A \\
& Refined  & \textbf{4.908} & \textbf{4.910} & \textbf{4.766} \\
\midrule
\multirow{2}{*}{Chinese} 
& Initial  & 4.855 & 3.603 & N/A \\
& Refined  & \textbf{4.995} & \textbf{4.926} & \textbf{4.865} \\
\bottomrule
\end{tabular}
\end{adjustbox}
\caption{Refinement evaluation results (RQ) in Korean and Chinese. Scores are based on a 5-point Likert scale averaged over LLM and human raters. Detailed are included in Appendix~\ref{appendix:e.1}}
\label{tab:refine-results}
\end{table*}

\paragraph{Dialogue Quality (DQ)}
We assess multi-turn dialogues along six criteria adapted from~\cite{kim2023soda,li2023normdial}. Our evaluation criteria are detailed in Table~\ref{tab:dq-criteria}. LLMs and human annotators independently conduct evaluations.

\paragraph{Generalization Quality (GQ)}
We fine-tune four language models on norm-adherent dialogues from three datasets. For training, we use 1,200 English and Chinese instances from our dataset. In addition, we use 1,265 English and 1,116 Chinese instances from \textsc{NormDial}, as well as 1,200 randomly sampled English dialogues from \textsc{SODA}. Models are evaluated on \textsc{DailyDialog} (English)~\cite{li2017dailydialog} and \textsc{LCCC} (Chinese)~\cite{wang2022largescalechineseshorttextconversation}, where each is prompted to generate five-turn continuations given benchmark dialogue contexts. We conduct A/B preference testing and human evaluation, with judgments based on social appropriateness, fluency, and overall response quality.

\section{Results}
\label{Sec:5}
We present results along the three evaluation axes defined in Section~\ref{Sec:4}. Also, we further analyze the impact of our \textit{Violation-to-Resolution} (V2R) modeling. Our framework consistently improves linguistic fluency, pragmatic coherence, and social appropriateness across all settings.

\subsection{Refinement Quality (RQ)}
\label{Sec:5.1}
To assess the effectiveness of exemplar-based refinement, we compare scenario-situation pairs before and after refinement in two low-resource languages: Korean and Chinese. As shown in Table~\ref{tab:refine-results}, refinement consistently improves all evaluation dimensions across both languages. In Korean, linguistic quality improves substantially, accompanied by gains in norm alignment and semantic fidelity. Similar patterns are observed in Chinese, with a +1.32 increase in linguistic quality and near-ceiling norm alignment scores.

The refinement process is repeated until the model output is no longer selected for revision by the LLM. On average, each instance undergoes 1.2 rounds of refinement. These findings underscore the utility of our approach in enhancing fluency, coherence, and sociocultural adequacy for low-resource language generation.

To test generalizability beyond East Asian typologies, we conducted pilot refinement experiments in Malay and Urdu, two pragmatically distinct languages. Using the same evaluation setup as in Table~\ref{tab:refine-results}, the observed improvements were consistent with those from our main experiments, supporting the robustness of our approach. Detailed results of these pilot studies are provided in Appendix~\ref{appendix:f.2}, which further illustrate the adaptability of our refinement strategy across diverse linguistic and cultural contexts.

\subsection{Dialogue Quality (DQ)}
\label{Sec:5.2}
We evaluated dialogue quality across six dimensions using both LLM- and human-based scoring. Criteria are summarized in Table~\ref{tab:dq-criteria}. For conciseness, we abbreviate the last three dimensions—emotional appropriateness, social norm appropriateness, and scenario–dialogue coherence—as \textit{Emo. Approp.}, \textit{Norm Approp.}, and \textit{Scenario Coh.}, respectively, throughout this section.

Table~\ref{tab:dqa_results} presents the average scores obtained from LLM-based evaluation for Adherence, V2R, and Violation scenarios in Korean, Chineses, and English. Dialogues from the \textit{Adherence} and \textit{V2R} categories consistently achieved high ratings (avg.\ $>$ 4.9), especially for consistency, emotional appropriateness, and scenario coherence. \textit{V2R} dialogues slightly outperformed others in emotional appropriateness and scenario–dialogue alignment, highlighting the framework's strength in modeling socially complex, repair-driven interactions. In contrast, Violation dialogues received lower scores in naturalness and emotional appropriateness across all languages, reflecting their design to capture socially inappropriate interactions.

\paragraph{Human Evaluation.}
To validate the robustness of our LLM-based assessments and to examine generation quality in low-resource settings, we conducted a parallel human evaluation in Korean and Chinese. 
As shown in Table~\ref{tab:dqa-human}, human ratings exhibit patterns highly consistent with LLM scores reported in Table~\ref{tab:dqa_results}. 
We further quantified the alignment between LLM and human evaluations using Pearson correlation. Results indicate strong agreement across both languages, with coefficients of $r = 0.928$ (Korean) and $r = 0.945$ (Chinese). These findings confirm the reliability of our automatic evaluation protocol and support the validity of the conclusions drawn from it.

\begin{table}[t]
\centering
\small
\begin{adjustbox}{max width=\linewidth}
\begin{tabular}{llccc}
\toprule
\textbf{Language} & \textbf{Criterion} & \textbf{Adherence} & \textbf{V2R} & \textbf{Violation} \\
\midrule
\multirow{6}{*}{Korean} 
& Consistency                  & \textbf{4.978}     & \textbf{4.978}     & 2.594 \\
& Naturalness                 & \textbf{5.000}     & 4.998     & 4.757 \\
& Relevance                   & 4.996     & \textbf{5.000}     & 4.996 \\
& Emo. Approp.     & 4.999     & \textbf{5.000}     & 3.375 \\
& Norm Approp. & \textbf{4.707}     & 3.816     & 1.613 \\
& Scenario Coh. & 4.988     & \textbf{5.000}     & 4.965 \\
\midrule
\multirow{6}{*}{Chinese} 
& Consistency                  & \textbf{5.000}     & 4.952     & 1.662 \\
& Naturalness                 & \textbf{4.432}     & 4.361     & 2.700 \\
& Relevance                   & 4.987     & \textbf{5.000}     & 4.950 \\
& Emo. Approp.     & \textbf{5.000}     & 4.987     & 1.918 \\
& Norm Approp. & \textbf{4.980}     & 3.528     & 1.216 \\
& Scenario Coh. & 4.896     & \textbf{5.000}     & 4.811 \\
\midrule
\multirow{6}{*}{English}
& Consistency & \textbf{5.000} & 4.947 & 1.665 \\
& Naturalness & \textbf{4.900} & 4.623 & 3.381 \\
& Relevance & \textbf{5.000} & \textbf{5.000} & 4.842 \\
& Emo. Approp. & \textbf{5.000} & 4.992 & 2.589 \\
& Norm Approp. & \textbf{4.982} & 3.241 & 1.186 \\
& Scenario Coh. & \textbf{4.994} & 4.932 & 4.801 \\
\bottomrule
\end{tabular}
\end{adjustbox}
\caption{Dialogue quality scores across six dimensions, evaluated on three scenario types (Adherence, \textit{Violation-to-Resolution} (V2R), Violation) in three languages.}
\label{tab:dqa_results}
\end{table}

\begin{table}[t]
\centering
\small
\begin{adjustbox}{max width=\linewidth}
\begin{tabular}{llccc}
\toprule
\textbf{Language} & \textbf{Criterion} & \textbf{Adherence} & \textbf{V2R} & \textbf{Violation} \\
\midrule
\multirow{6}{*}{Korean} 
& Consistency  & 4.500 & \textbf{5.000} & 2.500 \\
& Naturalness  & 4.250 & \textbf{4.625} & 4.125 \\
& Relevance    & \textbf{4.750} & \textbf{4.750} & 4.250 \\
& Emo. Approp. & \textbf{4.625} & 4.500 & 3.250 \\
& Norm Approp. & \textbf{4.500} & 4.375 & 1.375 \\
& Scenario Coh. & \textbf{4.750} & 4.625 & 4.500 \\
\midrule
\multirow{6}{*}{Chinese} 
& Consistency  & \textbf{4.500} & 4.375 & 1.500 \\
& Naturalness  & \textbf{4.625} & 4.500 & 3.125 \\
& Relevance    & \textbf{5.000} & 4.750 & 4.500 \\
& Emo. Approp. & 4.500 & \textbf{5.000} & 1.375 \\
& Norm Approp. & \textbf{4.750} & 4.500 & 1.500 \\
& Scenario Coh. & \textbf{4.750} & 4.625 & 4.125 \\
\bottomrule
\end{tabular}
\end{adjustbox}
\caption{Human evaluation results across dialogue types (Adherence, \textit{Violation-to-Resolution} (V2R), Violation) and six dimensions. Scores are based on a 5-point Likert scale.}
\label{tab:dqa-human}
\end{table}

\subsection{Generalization Quality (GQ)}
\label{Sec:5.3}
Automatic evaluation results are summarized in Table~\ref{tab:ab-results}. Across all models and languages, our dataset yields consistently higher preference than \textsc{NormDial} and \textsc{SODA}. For instance, GPT-4o-mini was preferred in 65\% (English) and 75\% (Chinese) of cases over \textsc{NormDial}, and in 65\% (English) over \textsc{SODA}. Larger models such as Qwen-2.5-32B showed similar trends, with the strongest preference observed in Chinese. To validate these findings, we conducted blind human evaluations in Korean and Chinese (Table~\ref{tab:human-eval}). Native speakers favored our dataset in 68\% (Korean) and 77\% (Chinese) of cases, closely matching LLM preferences. These results suggest that models trained on our dataset generalize better across languages and domains, generating more socially appropriate and contextually aligned responses.

We note that the untuned model occasionally produced concise, direct responses that some evaluators preferred in contexts requiring rapid apologies without nuanced emotional transitions. However, such cases were limited, and overall, our dataset achieved more than double the preference rate of the baseline. This result suggests that exemplar-guided refinement not only enhances overall quality but also provides a flexible framework for fine-grained control of response style and emotional expression, which we plan to investigate in future work.

\begin{table}[t]
\centering
\small
\begin{adjustbox}{max width=\linewidth}
\begin{tabular}{llcc}
\toprule
\textbf{Model} & \textbf{Language} & \textbf{Ours vs. NormDial} & \textbf{Ours vs. SODA} \\
\midrule
\multirow{2}{*}{GPT-4o-mini} & English & \textbf{65\%} & \textbf{65\%} \\
                             & Chinese & \textbf{75\%} & N/A \\
\midrule
\multirow{1}{*}{LLaMA-3-8B}  & English & \textbf{65\%} & \textbf{59\%} \\
\midrule
\multirow{2}{*}{Qwen-2.5-14B} & English & \textbf{51\%} & \textbf{61\%} \\
                              & Chinese & \textbf{71\%} & N/A \\
\midrule
\multirow{2}{*}{Qwen-2.5-32B} & English & \textbf{56\%} & \textbf{62\%} \\
                              & Chinese & \textbf{79\%} & N/A \\
\bottomrule
\end{tabular}
\end{adjustbox}
\caption{A/B test results comparing preference for models trained on our dataset versus NormDial and SODA. Each value represents the percentage of times responses from models trained on \textit{our dataset} were preferred by annotators.}
\label{tab:ab-results}
\end{table}

\begin{table}[t]
\centering
\small
\begin{adjustbox}{max width=\linewidth}
\begin{tabular}{llccc}
\toprule
\textbf{Language} & \textbf{Ours} & \textbf{Untuned GPT-4o-mini} & \textbf{NormDial} \\
\midrule
\multirow{1}{*}{Korean} & \textbf{68\%} & 32\% & N/A \\
\midrule
\multirow{1}{*}{Chinese} & \textbf{77\%} & 5\% & 18\% \\
\bottomrule
\end{tabular}
\end{adjustbox}
\caption{Human preference results comparing our dataset with untuned GPT-4o-mini and NormDial. Evaluations were conducted under blind conditions with native speakers. Korean evaluation compares against untuned GPT-4o-mini, while Chinese includes baseline.}
\label{tab:human-eval}
\end{table}

\subsection{Effect of Violation-to-Resolution}
\label{Sec:5.4}
To assess the impact of \textit{Violation-to-Resolution} (V2R) training on norm-sensitive generation, we conduct a focused comparison using \textsc{ProsocialDialog}~\cite{kim2022prosocialdialog}, a benchmark for ethically challenging scenarios. We fine-tune two GPT-4o-mini models under comparable conditions: one on the full NormDial dataset and one on an equal-sized subset of our data, including three types of our datasets. Both models are prompted with 100 norm-violating contexts, each requiring a five-turn continuation.

Blind A/B human evaluations show that the V2R-augmented model is preferred in 82\% of cases (Table~\ref{tab:v2r-results}), with annotators consistently favoring its empathy, contextual fit, and ability to model norm repair. These findings underscore the utility of V2R as a training signal for enhancing pragmatic competence in ethically sensitive dialogue and support its integration into norm-grounded generation frameworks.

\section{Discussion}
\label{Sec:6}

\paragraph{Limitations of Prompt-Based Generation in Low-Resource Contexts.}
Our refinement framework is motivated in part by the limitations of prior prompt-based approaches to social norm generation~\cite{li2023normdial, zhan-etal-2024-renovi}. These methods typically rely on static prompts with minimal norm signals, placing the burden of generation entirely on the language model. While this approach may yield fluent and contextually appropriate responses in high-resource languages such as English, it often results in pragmatic failures in low-resource settings like Korean and Chinese.

In Korean, generated dialogues frequently exhibit lexical redundancy (e.g., repeated expressions) and tone mismatches (e.g., informal apologies in formal contexts). In Chinese, issues include unnatural phrasing, exaggerated emotional responses, repetitive honorifics, and inconsistent tone from register mixing. These limitations underscore the difficulty of capturing fine-grained sociocultural norms through prompt-only methods.

To mitigate these issues, we introduce a refinement framework to improve fluency and norm alignment in low-resource settings. Additional examples appear in Appendix~\ref{appendix:f.1}.

\paragraph{Early Refinement for Sociocultural Alignment.}
As shown in Section~\ref{Sec:5.1}, even a small set of high-quality exemplars at this stage improves fluency and norm alignment in low-resource languages. Prior to refinement, we conducted a comparative analysis between model-generated outputs and native-authored revisions, which revealed recurring issues such as overuse of formulaic expressions, limited gesture variety, register–context mismatch, and weakened hierarchical cues. A key insight from this analysis is the distinction between surface accuracy and cultural appropriateness: model outputs may be grammatically and semantically correct, yet fail to include ritualistic or affective elements (e.g., apologies, condolences) that are essential to pragmatic expectations. These shortcomings were particularly salient in low-resource settings and are illustrated in Appendix~\ref{appendix:f.1}.

When optimized early in the generation pipeline, scenario–situation pairs provide strong social cues that guide the construction of coherent and culturally aligned dialogues. This early-stage refinement approach also aligns with recent works in controllable generation and structured planning~\cite{moryossef2019step, rashkin2021increasing}, which emphasize the role of explicit context modeling in coherence and goal alignment. Since social norms are inherently entangled with relational roles, power dynamics, and situational contexts, refining situational priors ensures that pragmatic and culturally appropriate behaviors emerge naturally. Our refinement stage thus functions not as a post hoc correction layer, but as a core mechanism for embedding sociocultural alignment into the generative process.

\begin{table}[t]
\centering
\small
\begin{tabular}{lc}
\toprule
\textbf{Training Data} & \textbf{Preference (\%)} \\
\midrule
Ours (with V2R)        & \textbf{82} \\
NormDial               & 18 \\
\bottomrule
\end{tabular}
\caption{Human preference results on \textsc{ProsocialDialog}. Models trained on our dataset with V2R significantly outperformed those trained on NormDial.}
\label{tab:v2r-results}
\end{table}

\begin{table}[t]
\centering
\small
\begin{adjustbox}{max width=\linewidth}
\begin{tabular}{llccc}
\toprule
\textbf{Category} & \textbf{Strategy / Sequence} & \textbf{English} & \textbf{Chinese} & \textbf{Korean} \\
\midrule
\multirow{5}{*}{\textbf{Strategy (\%)}} 
& Apology (A)      & 98.7 & 93.0 & 92.6 \\
& Explanation (X)  & 91.1 & 88.1 & 86.3 \\
& Empathy (E)      & 90.6 & 62.0 & 89.2 \\
& Compensation (C) & 82.8 & 79.3 & 72.4 \\
& Humor (H)        & 12.6 & 7.9  & 12.0 \\
\midrule
\multirow{3}{*}{\textbf{Top Sequence}} 
& Sequence & X $\rightarrow$ A $\rightarrow$ C & A $\rightarrow$ X $\rightarrow$ C & E $\rightarrow$ A $\rightarrow$ X \\
& Frequency (\%) & 33 & 29 & 32 \\
\bottomrule
\end{tabular}
\end{adjustbox}
\caption{Strategy usage rates and most frequent recovery sequences in V2R dialogues across American English, Chinese, and Korean.}
\label{tab:v2r-strategies}
\end{table}

\paragraph{Qualitative Insights into Cross-Linguistic Repair Strategies}
Our qualitative analysis of Violation-to-Resolution (V2R) dialogues shows that the proposed framework effectively models both universal and culture-specific repair strategies. As summarized in Table~\ref{tab:v2r-strategies}, Apology and Explanation dominate across English, Chinese, and Korean. Still, sequencing patterns diverge in culturally meaningful ways: English dialogues most often follow Explanation → Apology → Compensation, Chinese dialogues Apology → Explanation → Compensation, and Korean dialogues Empathy → Apology → Explanation. These results, informed by established taxonomies in apology and politeness research~\cite{radu2019empathy, zhang2024impact}, validate the framework’s ability to capture nuanced cross-linguistic variation.
Notably, over 85\% of V2R dialogues employed multi-step recovery, confirming that socially coherent repair rarely occurs through a single act. By integrating exemplar-guided refinement, our approach models these layered dynamics, bridging computational dialogue generation with sociolinguistic insights. These findings highlight the dual value of the V2R paradigm: providing a scalable schema for realistic dialogue repair and serving as a diagnostic lens for cultural variation. This adaptability also points to promising directions for extending NormGenesis to typologically diverse languages (e.g., Arabic, Swahili, Hindi), where divergent pragmatic systems pose additional challenges.

\section{Conclusion}
\label{Sec:7}
We present NormGenesis, a multicultural framework for generating and refining socially grounded dialogues in English, Chinese, and Korean. To address cultural and pragmatic limitations of existing dialogue systems, particularly in low-resource settings, we introduce an exemplar-based iterative refinement applied at the scenario-situation level. This upstream refinement design enables early alignment with linguistic, emotional, and sociocultural expectations, reducing generation errors before full dialogue synthesis.
We further propose a novel dialogue type, \textit{Violation-to-Resolution} (V2R), which models the recovery process following norm violations through repair strategies. V2R facilitates more realistic and context-sensitive modeling of social interaction dynamics. Our experimental results show that V2R not only improves pragmatic competence in ethically sensitive scenarios but also enhances generalization across languages and domains.
Through comprehensive human and LLM-based evaluations, we demonstrate that NormGenesis consistently outperforms existing datasets such as \textsc{NormDial} and \textsc{SODA} across multiple dimensions, including norm alignment, emotional coherence, and repair quality. By integrating linguistically and culturally diverse norms, fine-grained turn-level annotations, and structured refinement, NormGenesis provides a scalable and robust foundation for norm-aware dialogue modeling in multilingual and multicultural contexts.

\section*{Limitations}
While NormGenesis achieves strong performance across linguistic, emotional, and social dimensions, we acknowledge several limitations that open key directions for future work.
\paragraph{Language Coverage.}
Our framework currently supports only English, Chinese, and Korean. While these languages span a spectrum of resource availability and cultural characteristics, the framework does not address the full diversity of global languages and interactional norms. Expanding NormGenesis to additional languages—particularly those with limited computational resources or distinct social conventions (e.g., Arabic, Hindi, Swahili)—remains a key avenue for future work. 

To test generalizability beyond East Asian typologies, we conducted pilot refinement in Malay and Urdu, two pragmatically distinct languages. The results closely aligned with those from our main experiments, demonstrating the framework’s capacity to generalize to typologically and culturally diverse settings. Detailed results are presented in Appendix~\ref{appendix:f.2}. Building on this, we plan to extend NormGenesis to further low-resource languages, including Arabic and Swahili.
\paragraph{Exemplar Scalability.}
The iterative refinement process relies on a small number of manually revised exemplars. Although this approach is more scalable than full human annotation, scaling to a large number of new norms or domains could still be resource-intensive. To mitigate this, future work will explore active learning for efficient exemplar selection, a structured norm-centric repository for retrieval-based reuse, and clustering of culturally aligned regions (e.g., via World Values Survey) to enable exemplar transfer. These strategies aim to improve coverage and ensure scalable, high-quality refinement.
\paragraph{Evaluation and Subjectivity.}
Evaluating social norm adherence and conversational appropriateness inevitably involves subjectivity and cultural bias, especially across diverse sociolinguistic contexts. To mitigate this, we adopted three safeguards: (1) detailed rubrics assessing norm alignment, fluency, and emotional appropriateness (Appendix E); (2) native speaker annotators with cultural expertise; and (3) cross-review by multiple experts to offset exemplar-induced bias. While these measures and high inter-annotator agreement enhance reliability, further reducing cultural subjectivity remains an open challenge. Future research could explore culturally calibrated evaluation protocols or leverage LLM-based evaluators fine-tuned on localized criteria.
\paragraph{Norm Evolution.}
Social norms are dynamic and shift across time, communities, and platforms. Our taxonomy provides a structured but time-bounded snapshot. Systematically tracking and modeling the evolution of norms over time and across social contexts will be important for adaptive and future-proof dialogue systems. Future work should systematically track norm shifts using longitudinal corpora or real-time social data, enabling adaptive norm modeling for evolving conversational environments.

\section*{Ethical Considerations}
NormGenesis aims to advance the development of culturally adaptive and socially competent dialogue agents by modeling nuanced social norms across English, Chinese, and Korean. However, several ethical considerations warrant discussion.
\paragraph{Intended Use and Misuse.}
Our dataset is designed for training dialogue systems to generate socially appropriate, norm-aware responses. As with any dataset containing norm violations and repair strategies, there is a risk that malicious users could exploit the resource to train agents that generate inappropriate or harmful utterances. We urge the community to use the dataset solely for prosocial, culturally sensitive, and norm-aligned conversational AI research.
\paragraph{Cultural Scope and Generalizability.}
NormGenesis reflects culturally salient behaviors as of the time of data collection. Despite extensive native speaker review and expert refinement, social norms are inherently dynamic and context-dependent. Caution is advised when applying the resource to new languages, regions, or changing societal contexts, as some outputs may not generalize beyond the represented cultures.
\paragraph{Annotation Subjectivity and Bias.}
All subnorms and dialogues are reviewed or annotated by cultural experts and native speakers, but the interpretation of social appropriateness and emotional tone involves subjective judgment. While inter-annotator agreement is high, some bias may persist, especially for edge cases or rapidly changing norms. Broadening annotator diversity and incorporating community feedback may mitigate these effects.
\paragraph{Dataset Balance and Representation.}
NormGenesis includes a diverse range of norm-adhering, violating, and Violation-to-Resolution (V2R) dialogues. However, its scenario coverage is not exhaustive and may reflect existing cultural, demographic, or linguistic biases. Supplementing NormGenesis with additional resources is encouraged to ensure robust and contextually sensitive conversational agents.
\paragraph{Potential for Negative Outcomes.}
While the V2R paradigm models constructive responses to norm violations, dialogue agents trained on these data should not be used for critical decision-making or sensitive applications (e.g., counseling, legal advice) without careful human oversight. The framework is intended to support research and development in social dialogue modeling, not to replace professional judgment.
All code, data, and annotation guidelines will be released publicly upon acceptance, promoting transparency, reproducibility, and responsible community use.

\section*{Acknowledgments} 
This research was supported by the MSIT (Ministry of Science and ICT), Korea, under the ITRC (Information Technology Research Center) support program (IITP-2025-RS-2020-II201789), and the Artificial Intelligence Convergence Innovation Human Resources Development (IITP-2025-RS-2023-00254592) supervised by the IITP (Institute for Information \& Communications Technology Planning \& Evaluation).

\appendix

\section{Dataset Overview}
\label{appendix:a}
\subsection{Dataset Composition Summary}
Our dataset includes multicultural social norm dialogues across 12 categories: \textit{Apology, Compliment, Condolence, Criticism, Empathy, Greeting, Leave-taking, Persuasion, Request, Respect, Responding to Compliments, Thanks}. For each category, we define 10 subnorms per language (English, Korean, Chinese), resulting in 120 subnorms per language.

\begin{itemize}
    \item \textbf{Total Subnorms}: 360 (120 per language)
    \item \textbf{Total Scenario--Situation Pairs}: 10,800 (3 types $\times$ 3 languages $\times$ 1,200 each)
    \item \textbf{Total Dialogues}: 10,800 (1 per instance)
    \item \textbf{Total Average Turn}: 11.91 Turn.
\end{itemize}

\subsection{Norm-Type Statistics}
To provide an overview of the social norm taxonomy introduced in this study, Table~\ref{appendix:a.2:adhere},~\ref{appendix:a.2:v2r}, and~\ref{appendix:a.2:violation} present detailed summary statistics for the Adherence, Violation-to-Resolution (V2R), and Violation categories, respectively. For brevity, column headers are abbreviated as follows: Lang = Target Language, Cat = Norm Category, Sub = Subnorm, Scen = Scenario, Situ = Situation, Dial = Dialogue, AvgT = Average Turn Count per dialogue. These tables summarize the scale and structure of our dataset across all categories and languages.

\begin{table}[t]
\centering
\textbf{Adherence}
\vspace{0.5em}
\begin{adjustbox}{max width=\linewidth}
\begin{tabular}{l|c|c|c|c|c|c}
\toprule
\textbf{Lang} & \textbf{Cat} & \textbf{Sub} & \textbf{Scen} & \textbf{Situ} & \textbf{Dial} & \textbf{AvgT} \\
\midrule
EN & 12 & 120 & 1200 & 1200 & 1200 & 10.56 \\
KR & 12 & 120 & 1200 & 1200 & 1200 & 10.21 \\
ZH & 12 & 120 & 1200 & 1200 & 1200 & 10.74 \\
\bottomrule
\end{tabular}
\end{adjustbox}
\caption{Dataset statistics for Adherence}
\label{appendix:a.2:adhere}
\end{table}

\begin{table}[t]
\centering
\textbf{Violation-to-Resolution (V2R)}
\vspace{0.5em}
\begin{adjustbox}{max width=\linewidth}
\begin{tabular}{l|c|c|c|c|c|c}
\toprule
\textbf{Lang} & \textbf{Cat} & \textbf{Sub} & \textbf{Scen} & \textbf{Situ} & \textbf{Dial} & \textbf{AvgT} \\
\midrule
EN & 12 & 120 & 1200 & 1200 & 1200 & 16.06 \\
KR & 12 & 120 & 1200 & 1200 & 1200 & 12.56 \\
ZH & 12 & 120 & 1200 & 1200 & 1200 & 13.49 \\
\bottomrule
\end{tabular}
\end{adjustbox}
\caption{Dataset statistics for violation-to-resolution (V2R)}
\label{appendix:a.2:v2r}
\end{table}

\begin{table}[t]
\centering
\textbf{Violation}
\vspace{0.5em}
\begin{adjustbox}{max width=\linewidth}
\begin{tabular}{l|c|c|c|c|c|c}
\toprule
\textbf{Lang} & \textbf{Cat} & \textbf{Sub} & \textbf{Scen} & \textbf{Situ} & \textbf{Dial} & \textbf{AvgT} \\
\midrule
EN & 12 & 120 & 1200 & 1200 & 1200 & 11.59 \\
KR & 12 & 120 & 1200 & 1200 & 1200 & 11.17 \\
ZH & 12 & 120 & 1200 & 1200 & 1200 & 10.82 \\
\bottomrule
\end{tabular}
\end{adjustbox}
\caption{Dataset statistics for violation}
\label{appendix:a.2:violation}
\end{table}

\subsection{Subnorm Coverage and Cultural Examples}
Each of the 12 norm categories comprises 10 culturally grounded subnorms per language, resulting in a total of 360 subnorm definitions with aligned examples across English, Korean, and Chinese. These examples serve as reference points for scenario generation and support culturally appropriate dialogue construction in each language. Full examples are provided in Appendix~\ref{appendix:c}.

\subsection{Instance Structure}
Each dialogue instance in the dataset is composed of the following stages:
\begin{itemize}
 \item \textbf{Norm Category \& Subnorm:} A high-level social norm category and its culturally grounded subnorm definition.
 \item \textbf{Scenario:} A brief 1--2 sentence description outlining the situational context in which the norm is relevant.
 \item \textbf{Situation:} A 3--5 sentence elaboration of the scenario that specifies tone, interpersonal relationship, and emotional cues to guide dialogue generation.
 \item \textbf{Dialogue:} A multi-turn conversation consisting of 5 to 15 turns that reflects the defined norm and situational context.
 \item \textbf{Annotations:} Turn-level labels for social norm adherence (e.g., Adherence, Violated, V2R) and speaker reactions (e.g., Apology, Empathy, Agreement), enabling fine-grained evaluation of social behavior and pragmatic intent.
 \end{itemize}
This structured format enables controlled generation and fine-grained annotation of socially grounded dialogues across multiple languages.

\subsection{Language Balance and Complexity Metric}
To ensure cultural and linguistic balance, each language is equally represented across all norm types. While average token length is commonly used to measure dialogue complexity, we omit it here due to tokenizer variations across models. Instead, we report average dialogue turns as a consistent and model-independent proxy for complexity.

\begin{table*}[t]
\centering
\small
\begin{tabularx}{\textwidth}{l|X}
\toprule
\textbf{Label} & \textbf{Description} \\
\midrule
\textbf{ACK} (Acknowledgment) & Explicit acknowledgment that the speaker has heard or understood the interlocutor’s statement. \\
\midrule
\textbf{AGR} (Agreement) & Expressing agreement or alignment with the other person’s opinion or position. \\
\midrule
\textbf{DIS} (Disagreement / Refusal) & Expressing disagreement or rejecting a suggestion. \\
\midrule
\textbf{APO} (Apology) & Expressing regret or remorse for one’s mistake or inconvenience caused. \\
\midrule
\textbf{THX} (Gratitude) & Expressing appreciation for help, kindness, or praise. \\
\midrule
\textbf{EMP} (Empathy / Support) & Emotionally validating or supporting the interlocutor’s feelings or situation. \\
\midrule
\textbf{JUS} (Justification) & Offering an explanation or excuse to justify one’s actions or mistakes. \\
\midrule
\textbf{SUG} (Suggestion / Advice) & Proposing a solution or sharing an opinion to help resolve an issue. \\
\midrule
\textbf{QUE} (Question / Clarification Request) & Asking a question to seek further explanation or information. \\
\midrule
\textbf{CRT} (Criticism) & Pointing out problems or expressing negative evaluations of the other person’s actions or statements. \\
\midrule
\textbf{N/A} (Not Applicable) & Used when the utterance does not clearly fall into any of the defined categories above. Typically applies to greetings, topic transitions, structural openers, or filler phrases. \\
\bottomrule
\end{tabularx}
\caption{Turn-level annotation labels and descriptions}
\label{tab:label-schema}
\end{table*}

\begin{table*}[t]
\centering
\begin{tabularx}{\textwidth}{l|X}
\toprule
\textbf{Section} & \textbf{Content} \\
\midrule
\textbf{Target Language} & \texttt{< Korean >} \\
\midrule
\textbf{Parameter} & \texttt{< WVS Responses >} \\
\midrule
\textbf{Instruction} & 
You are a culturally aware assistant with deep knowledge of Korean social norms and communication practices. Given a specific social norm category (e.g., Apology, Empathy), your task is to generate 10 Korean-specific conversational subnorms that reflect the core values and expectations found in Korean society. These values are derived from nationally representative survey data, including interpersonal relationships, formality, group harmony, respect for authority, and emotional expression.

Please follow these instructions when generating the subnorms:

1. Ensure that each subnorm reflects Korean cultural values and not generic or universal norms.

2. Specify the context in which the norm should be applied (e.g., school, workplace, with elders).

3. Include verbal evidence (i.e., example phrases in Korean) that would signal adherence to the subnorm in dialogue.

4. The subnorms should be actionable and observable in conversation. \\
\midrule
\textbf{Format} & 
Subnorm 1

Subnorm 2

...

Subnorm 10 \\
\bottomrule
\end{tabularx}
\caption{Prompt for korean subnorm generation.}
\label{tab:subnorm-korean}
\end{table*}

\begin{table*}[t]
\centering
\begin{tabularx}{\textwidth}{l|X}
\toprule
\textbf{Section} & \textbf{Content} \\
\midrule
\textbf{Target Language} & \texttt{< Chinese | English >} \\
\midrule
\textbf{Parameter} & \texttt{< Korean subnorm | Normdial subnorm >} \\
\midrule
\textbf{Instruction} & 
You are a culturally adaptive assistant with knowledge of conversational norms across multiple languages. Provided with a Korean subnorm reflecting a specific cultural value (e.g., how to apologize, express empathy, or show respect), your task is to generate a corresponding conversational subnorm in (Chinese | English) that matches the Korean subnorm in meaning and function.

Use the following instructions:

1. Ensure the subnorm reflects the target language’s cultural norms (i.e., Chinese or English), not just a literal translation of the Korean input.

2. Specify the context in which the norm should be applied (e.g., school, workplace, with elders).

3. Include verbal evidence (i.e., example phrases in Chinese or English) that would signal adherence to the subnorm in dialogue.

4. The subnorms should be actionable and observable in conversation. \\
\midrule
\textbf{Format} & 
Subnorm 1

Subnorm 2

...

Subnorm 10 \\
\bottomrule
\end{tabularx}
\caption{Prompt for chinese and english subnorm generation.}
\label{tab:subnorm-other}
\end{table*}

\begin{table*}[t]
\centering
\begin{tabularx}{\textwidth}{l|X}
\toprule
\textbf{Section} & \textbf{Content} \\
\midrule
\textbf{Target Language} & \texttt{< Korean | English | Chinese >} \\
\midrule
\textbf{Category} & \texttt{< Apology >} \\
\midrule
\textbf{Instruction} & 
Based on the above \texttt{Subnorm} for the given \texttt{Category}, generate 10 distinct and concise scenarios that could naturally occur in a similar social context within \texttt{Country} culture. Please ensure realism and cultural grounding. Use names and honorifics commonly used in the specified country. \\
\midrule
\textbf{Input Format} & 
\texttt{Category, Subnorm, Instruction, Type} \\
\midrule
\textbf{Output Format} & 
Scenario 1  

Scenario 2  

...  

Scenario 10 \\
\bottomrule
\end{tabularx}
\caption{Prompt for scenario generation.}
\label{tab:scenario-generation}
\end{table*}

\begin{table*}[t]
\centering
\begin{tabularx}{\textwidth}{l|X}
\toprule
\textbf{Section} & \textbf{Content} \\
\midrule
\textbf{Target Language} & \texttt{< Korean | English | Chinese >} \\
\midrule
\textbf{Category} & \texttt{< Apology >} \\
\midrule
\textbf{Instruction} & 
Generate a culturally plausible Situation in 3--5 sentences.  
1. Use realistic names and honorifics.  
2. Ensure emotional coherence.  
3. Depict through action/dialogue, not explanation. \\
\midrule
\textbf{Input Format} & 
\texttt{Category, Subnorm, Scenario, Instruction} \\
\midrule
\textbf{Output Format} & 
Situation 1 

Situation 2 

...  

Situation 10 \\
\bottomrule
\end{tabularx}
\caption{Prompt for situation elaboration.}
\label{tab:Situation}
\end{table*}

\begin{table*}[t]
\centering
\begin{tabularx}{\textwidth}{l|X}
\toprule
\textbf{Section} & \textbf{Content} \\
\midrule
\textbf{Instruction} & 
Given a naive Scenario--Situation pair and an expert-refined version, rewrite all other naive inputs to match the expert style. Do not shorten content. Ensure cultural richness and tone. Cover a wide range of settings and relationships. Avoid repetition. \\
\midrule
\textbf{Input Format} & 
\texttt{Category, Subnorm, 9 Naive Scenarios, 9 Naive Situations, Exemplar} \\
\midrule
\textbf{Output Format} & 
Rewritten Scenario 1  

Rewritten Situation 1  

...  

Rewritten Scenario 9  

Rewritten Situation 9 \\
\bottomrule
\end{tabularx}
\caption{Prompt for exemplar-based scenario--situation refinement.}
\label{tab:refinement}
\end{table*}

\begin{table*}[t]
\centering
\begin{tabularx}{\textwidth}{l|X}
\toprule
\textbf{Section} & \textbf{Content} \\
\midrule
\textbf{Input stages} & 
\texttt{Category, Subnorm, Scenario, Situation} \\
\midrule
\textbf{Instruction} & 
Generate a natural and realistic dialogue between two speakers reflecting the Scenario and Situation above. Write in English. Format each turn with speaker names. Dialogue should be 5--15 turns long. \\
\midrule
\textbf{Output Format} & 
\texttt{Name: line of dialogue}

\texttt{Name: line of dialogue} 

... \texttt{[END]} \\
\bottomrule
\end{tabularx}
\caption{Prompt for multi-turn dialogue generation.}
\label{tab:Dialogue}
\end{table*}

\begin{table*}[t]
\centering
\begin{tabularx}{\textwidth}{l|X}
\toprule
\textbf{Section} & \textbf{Content} \\
\midrule
\textbf{Instruction} & 
For each dialogue turn, label: (1) Norm-level adherence (Adherence / Violation / Not Relevant), (2) Speaker's reaction label (e.g., APO, ACK, AGR), and (3) Explanation for label. \\
\midrule
\textbf{Label Format} & 
\texttt{Role | Norm Label | Reaction Label | Explanation} \\
\midrule
\textbf{Reaction Labels} & 
APO: Apology, ACK: Acknowledgment, AGR: Agreement, DIS: Disagreement, THX: Thanks, EMP: Empathy, JUS: Justification, SUG: Suggestion, QUE: Question, CRT: Criticism, N/A: Not applicable \\
\bottomrule
\end{tabularx}
\caption{Prompt for turn-level annotation.}
\label{tab:annotation}
\end{table*}

\begin{figure*}[t] 
\centering
\includegraphics[width=\textwidth]{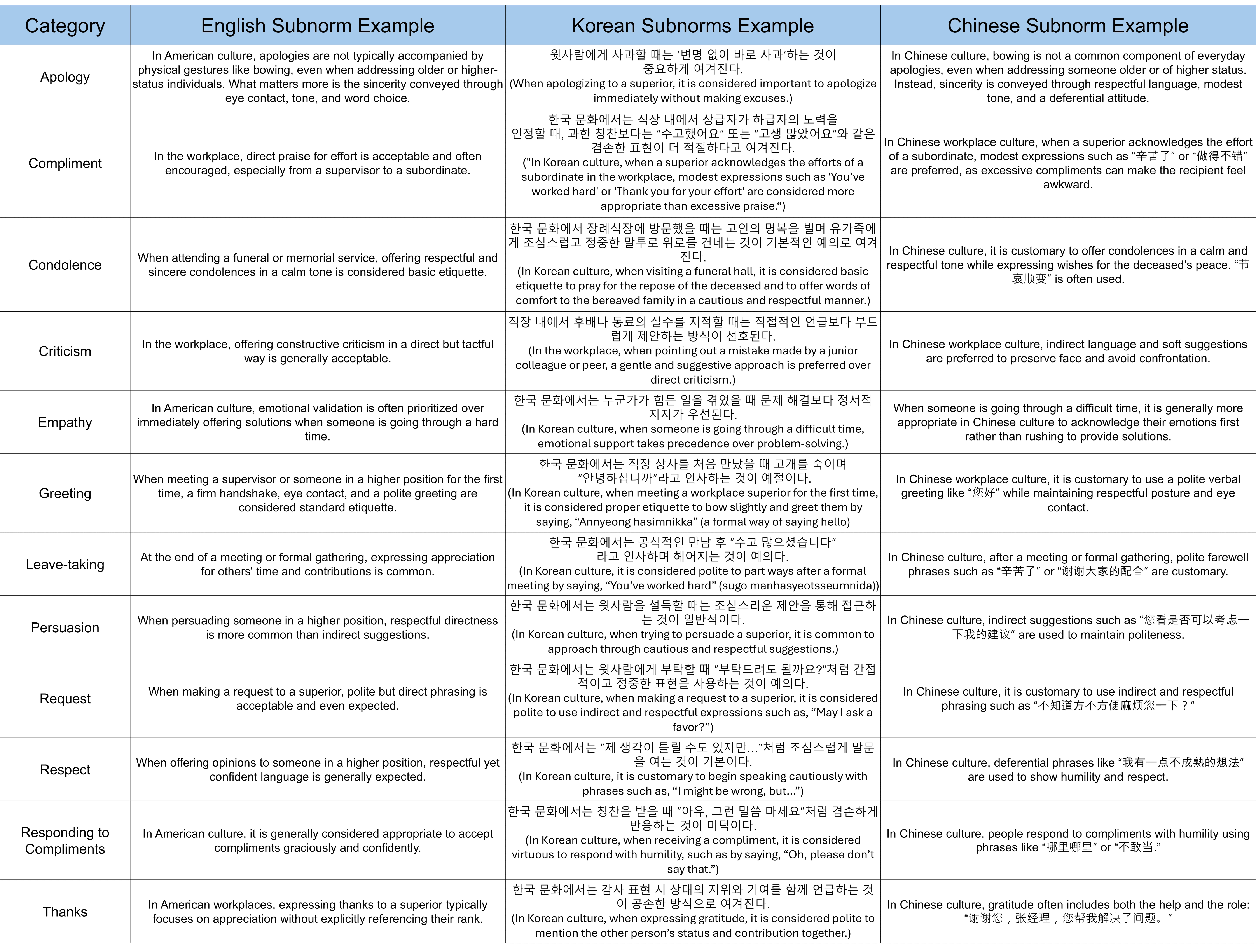} 
\caption{Subnorm examples}
\label{fig:subnorm}
\end{figure*}

\section{Annotation Schema}
\label{appendix:b}

This appendix describes the schema used for turn-level annotation of generated dialogues.
Each utterance is annotated with a communicative function label drawn from a predefined set of categories, facilitating fine-grained analysis of pragmatic intent and interactional structure across cultural contexts.

\subsection{Label Set Definition}

We employ a set of 11 functional dialogue act labels to annotate each turn, as summarized in Table~\ref{tab:label-schema}. These labels capture core social and communicative functions, including acknowledgment, apology, suggestion, justification, and other norm-relevant speaker intentions.

\subsection{Annotation Usage}
Each generated dialogue is annotated at the turn level. Given a dialogue $D = {u_1, ..., u_n}$ consisting of $n$ utterances, we assign a label $y_i \in \mathcal{Y}$ to each turn $u_i$ using an automatic annotation function $f_{\text{annotate}}$.
This annotation framework supports comparative analysis of pragmatic behavior across the following dimensions:
\begin{itemize}
 \item \textbf{Norm Type:} Adherence, Violation, Violation-to-Resolution (V2R)
 \item \textbf{Cultural Context:} English (EN), Korean (KR), Chinese (ZH)
 \item \textbf{Speaker Role:} e.g., subordinate vs. authority figure
 \end{itemize}
This enables a structured investigation of how social norms and communicative functions vary across languages and roles.

\subsection{Annotation Example}
Appendix~\ref{appendix:c.7} details the prompt formulations used to perform turn-level annotation.
For illustration, Figure~\ref{fig:Adherence-label} presents a turn-level annotation example for the “Adherence” category in English.

\section{Generation \& Refinement Prompt Templates}
\label{appendix:c}
This appendix presents the full set of prompt templates used throughout the multilingual dialogue generation and refinement pipeline. We first outline the overall algorithmic workflow, which defines the step-by-step procedures for constructing culturally grounded dialogues. We then provide task-specific prompt templates corresponding to each stage of the pipeline, designed to ensure consistency, linguistic fluency, and norm alignment across languages.

\subsection{Algorithm of Our Framework}
\label{appendix:c.1}
Algorithm~\ref{alg} outlines the complete pipeline used to construct our multilingual, norm-grounded dialogue dataset. The framework consists of four main stages:
\paragraph{Step 1: Social Norm Construction.}
For each language $l$ and social norm category $c$, we generate a set of subnorms $\mathcal{N}_{l,c}$ that encode fine-grained, culturally grounded expectations. These subnorms serve as foundational inputs for subsequent stages of scenario and dialogue generation.
\paragraph{Step 2: Scenario--Situation Generation.}
For each subnorm $n$ and dialogue type $t$ (e.g., Adherence, Violation, Violation-to-Resolution), we construct a scenario $S_{l,n,t}$ that outlines the relevant social context. This is followed by a situation $T_{l,n,t}$, which elaborates on tone, relationship, and emotional dynamics to constrain downstream dialogue construction.
\paragraph{Step 3: Exemplar-Based Iterative Refinement.}
To ensure cultural and pragmatic fidelity, an expert manually refines a single exemplar $(S_{l,n,t}, T_{l,n,t})$ per subnorm. This exemplar guides LLM-based refinement of structurally or semantically similar pairs. The refinement process is repeated until the quality score $Q_{l,n,t}$ exceeds a predefined threshold, ensuring consistency across instances.
\paragraph{Step 4: Dialogue Generation and Annotation.}
Each refined scenario--situation pair $(S_{l,n,t}, T_{l,n,t})$ is used to generate a multi-turn dialogue $D_{l,n,t}$, conditioned on the norm category, subnorm, scenario, and situation. All dialogue turns are annotated with norm and reaction labels, resulting in a fully labeled dataset $\mathcal{D}$ suitable for training and evaluation.
\paragraph{Summary.}
This structured pipeline enables the generation of culturally faithful, pragmatically coherent, and richly annotated dialogues. It supports multilingual benchmarking and serves as a scalable foundation for norm-aware dialogue modeling.

\subsection{Subnorm Generation Prompt}
\label{appendix:c.2}
To generate culturally grounded subnorms for the 12 norm categories defined in Section~\ref{Sec:3.1}, we construct 10 subnorms per category for each target language, yielding 360 subnorms in total. For Korean, where prior work on normative dialogue modeling is limited, we leverage sociocultural value indicators from the World Values Survey (WVS Wave 7, South Korea) as illustrated in Table~\ref{tab:subnorm-korean}.
For English and Chinese, we adopt the subnorm definitions from~\cite{li2023normdial} as semantic anchors. To ensure cross-cultural consistency, we design prompts that adapt these definitions to align with the Korean-derived subnorms. Table~\ref{tab:subnorm-other} illustrates representative prompt templates used for this alignment procedure. As a result, we design a total of 360 subnorms across the 12 categories and three languages, with illustrative examples shown in Figure~\ref{fig:subnorm}.
This protocol ensures cross-linguistic consistency while preserving cultural specificity and serves as the foundation for downstream scenario construction and dialogue generation.

\subsection{Scenario Generation Prompt}
\label{appendix:c.3}
To generate culturally grounded data, we first construct concise yet diverse scenarios aligned with a given subnorm within each social norm category. For each target language (English, Korean, Chinese), we prompt the model to produce 10 distinct and contextually appropriate scenarios per subnorm. The prompts are designed to ensure that the resulting scenarios are socially plausible and culturally relevant. As shown in Table~\ref{tab:scenario-generation}, these prompts provide structured guidance for consistent and culturally sensitive scenario generation.

\subsection{Situation Elaboration Prompt}
\label{appendix:c.4}
Building on each generated scenario, this prompt instructs the model to produce a realistic and emotionally coherent situation in 3--5 sentences. The generated text is expected to reflect culturally appropriate tone, interpersonal dynamics, and narrative plausibility. These situational descriptions serve as the contextual foundation for downstream dialogue generation. A representative example is provided in Table~\ref{tab:Situation}.

\subsection{Exemplar-Based Refinement Prompt}
\label{appendix:c.5}
To improve linguistic quality and pragmatic fidelity, particularly in low-resource languages, we employ a one-shot refinement strategy. The model is given an expert-curated Scenario–Situation pair as an exemplar and instructed to revise a batch of initial outputs to match the demonstrated level of cultural appropriateness, contextual richness, and fluency. Table~\ref{tab:refinement} illustrates the structure and usage of the refinement prompt.

\subsection{Multi-Turn Dialogue Generation Prompt}
\label{appendix:c.6}
Given a refined scenario and situation, this prompt guides the generation of a natural, coherent multi-turn dialogue that adheres to the specified social norm. Each dialogue comprises 5 to 15 turns and is expected to reflect appropriate cultural tone, relational dynamics, and norm-conforming behavior in a realistic conversational format. An illustrative example is provided in Table~\ref{tab:Dialogue}.

\subsection{Turn-Level Annotation Prompt}
\label{appendix:c.7}
To assess norm adherence and communicative function at the utterance level, this prompt guides the annotation of each dialogue turn with: 
(1) a norm label (Adherence, Violation, Not Relevant), 
(2) a communicative function tag (e.g., APO, ACK, THX), and 
(3) a brief justification.

This structured annotation facilitates consistent, turn-level analysis of social behaviors and pragmatic functions across cultures. An example is shown in Table~\ref{tab:annotation}.

\section{Generation \& Refinement Examples}
\label{appendix:d}
This appendix presents full examples of generated and refined outputs for each social norm type—\textit{Adherence}, \textit{Violation}, and \textit{Violation-to-Resolution (V2R)}—across English (EN), Korean (KR), and Chinese (ZH). For each case, we provide the subnorm definition, the Scenario--Situation pair before and after refinement, the corresponding dialogue, and representative turn-level annotations.

To facilitate narrative understanding and cross-cultural comparison, we include aligned figure references for each norm type and language. These visualizations illustrate the transformation process and demonstrate how refinement enhances linguistic fluency, pragmatic appropriateness, and sociocultural alignment. All examples are drawn from the \textit{Apology} category.

\subsection{Adherence Examples}
\begin{figure*}[t]
\centering
\includegraphics[width=\textwidth]{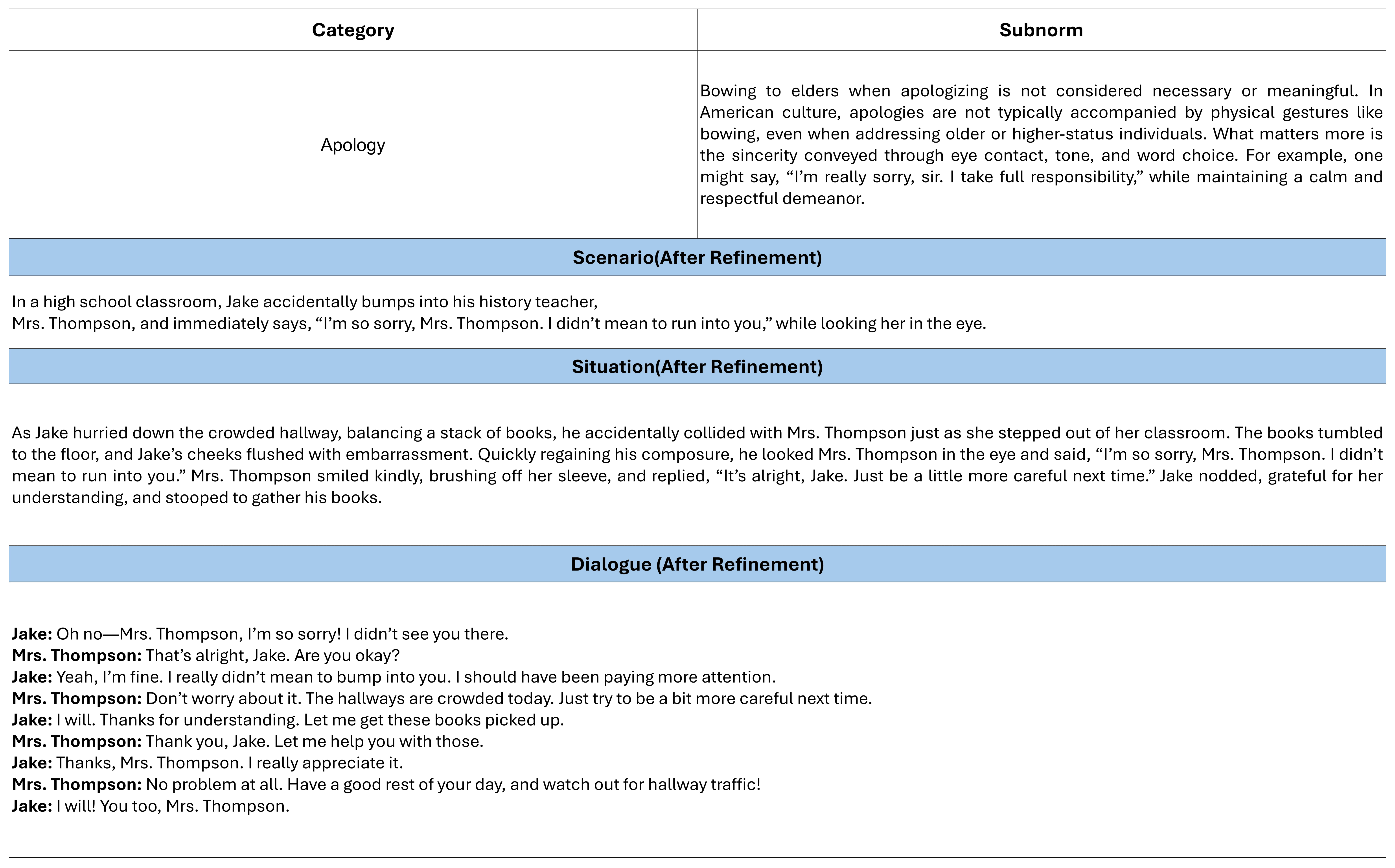}
\caption{Adherence norm example (EN)}
\label{fig:Adherence-EN}
\end{figure*}

\begin{figure*}[t]
\centering
\includegraphics[width=\textwidth]{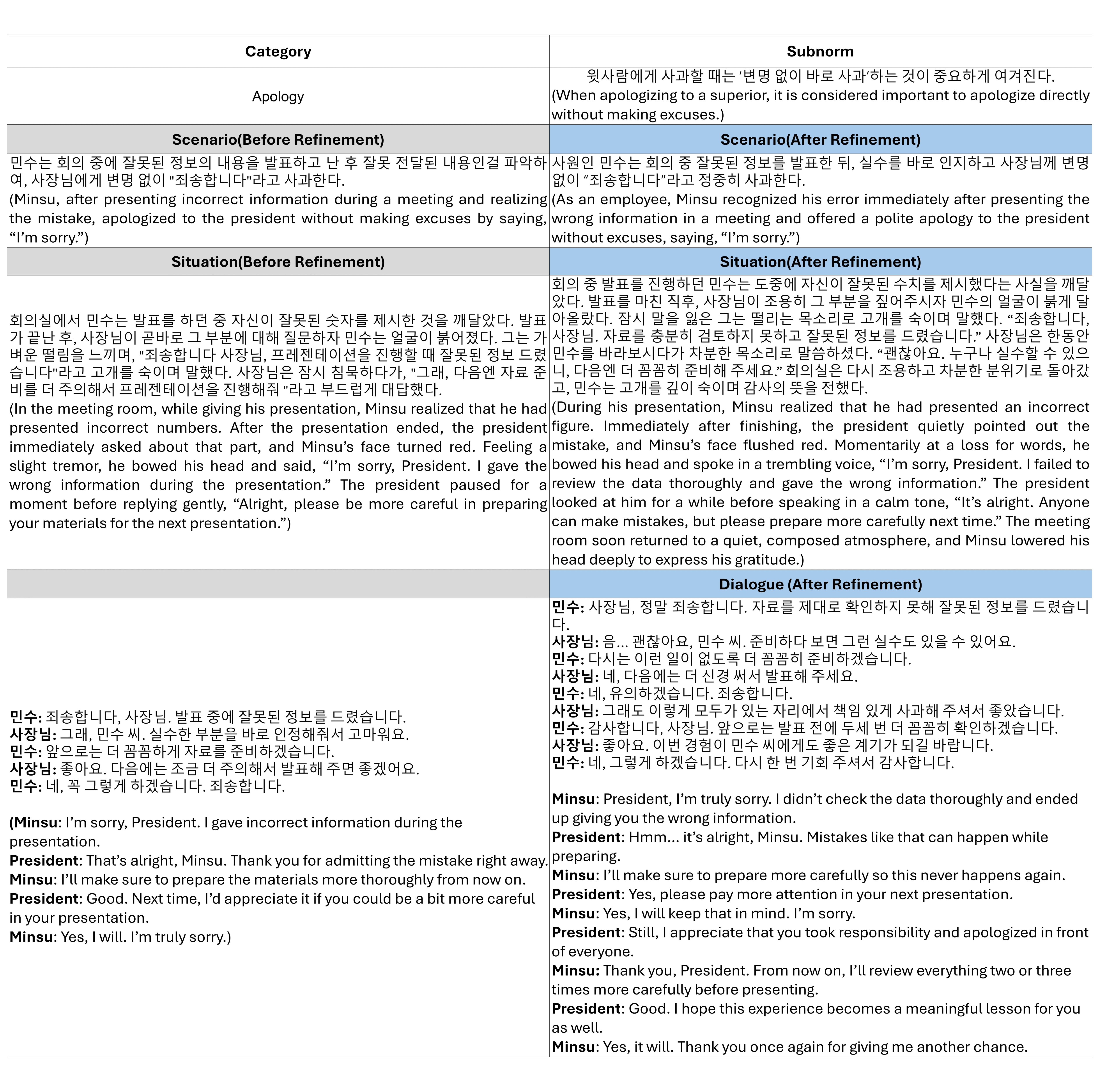}
\caption{Adherence norm example (KR)}
\label{fig:Adherence-KR}
\end{figure*}

\begin{figure*}[t]
\centering
\includegraphics[width=\textwidth]{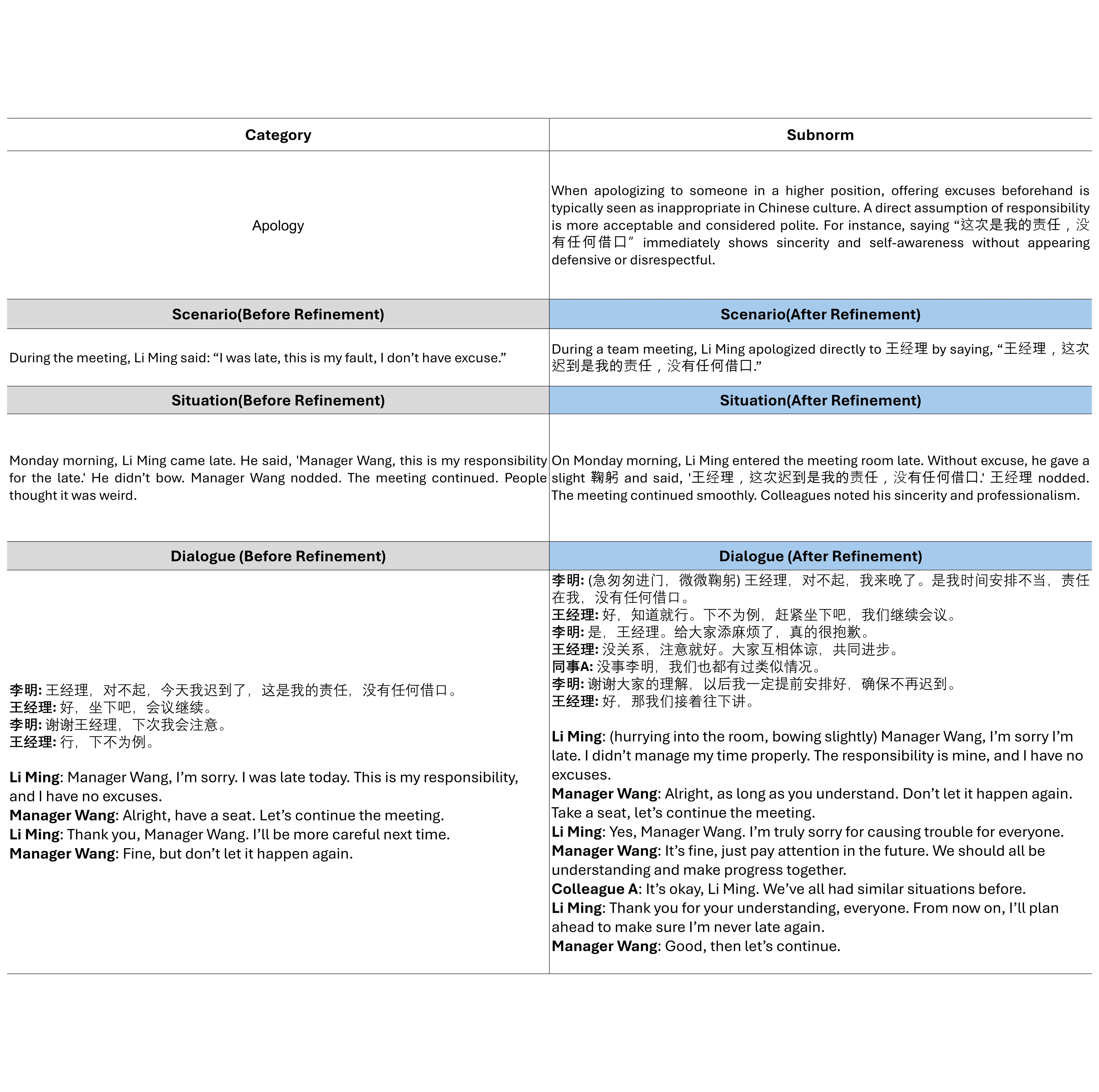}
\caption{Adherence norm example (ZH)}
\label{fig:Adherence-ZH}
\end{figure*}

\begin{figure*}[t]
\centering
\includegraphics[width=\textwidth]{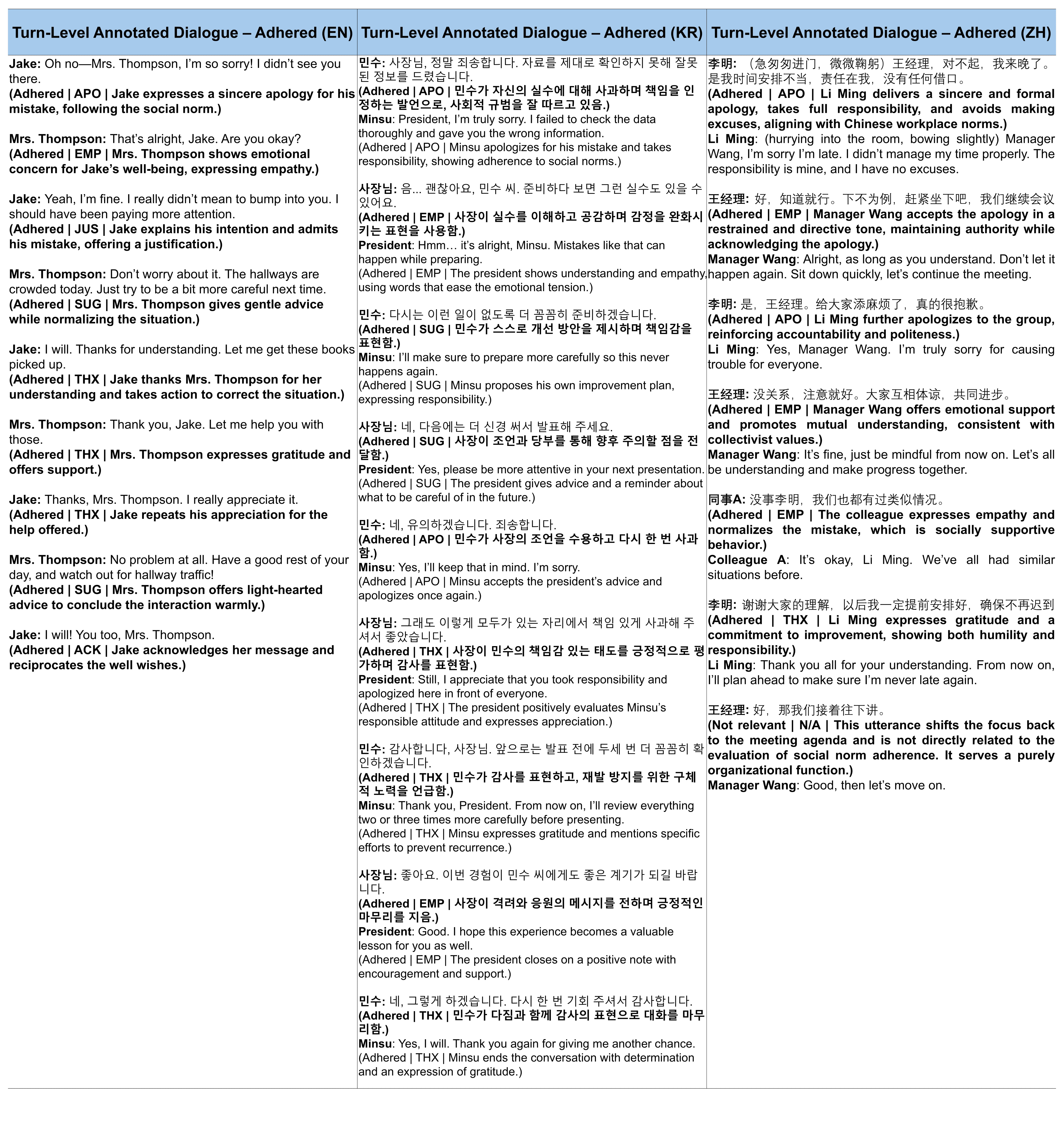}
\caption{Turn-level annotated dialogue example – Adherence}
\label{fig:Adherence-label}
\end{figure*}

This section presents example dialogues that adhere to social norms from the initial generation stage, across English, Korean, and Chinese. Although these dialogues already demonstrate norm-conforming behavior, we apply exemplar-based refinement to enhance fluency, tonal consistency, and cultural appropriateness.
In high-resource languages like English, model outputs tend to exhibit strong cohesion and emotional clarity even before refinement (Figure~\ref{fig:Adherence-EN}. The initial dialogue demonstrates contextual awareness and sincerity with minimal pragmatic inconsistencies. In contrast, the Korean and Chinese examples reveal more pronounced cultural deviations. For instance, Korean outputs often lack the deference and softened phrasing expected in hierarchical contexts (Figure~\ref{fig:Adherence-KR}, while Chinese dialogues may omit expressions of empathy or communal responsibility crucial in professional settings (Figure~\ref{fig:Adherence-ZH}.
Post-refinement, all examples show marked improvements in pragmatic subtlety, such as calibrated tone, culturally appropriate honorifics, and clearer interpersonal alignment, yielding more realistic and socially congruent conversations.
Turn-level annotation is subsequently applied to each dialogue, capturing both norm adherence and speaker intent at the utterance level. An annotated example illustrating this process is shown in Figure~\ref{fig:Adherence-label}.

\subsection{Violation-to-Resolution(V2R) Examples}
\label{appendix:d.2}
\begin{figure*}[t]
\centering
\includegraphics[width=\textwidth]{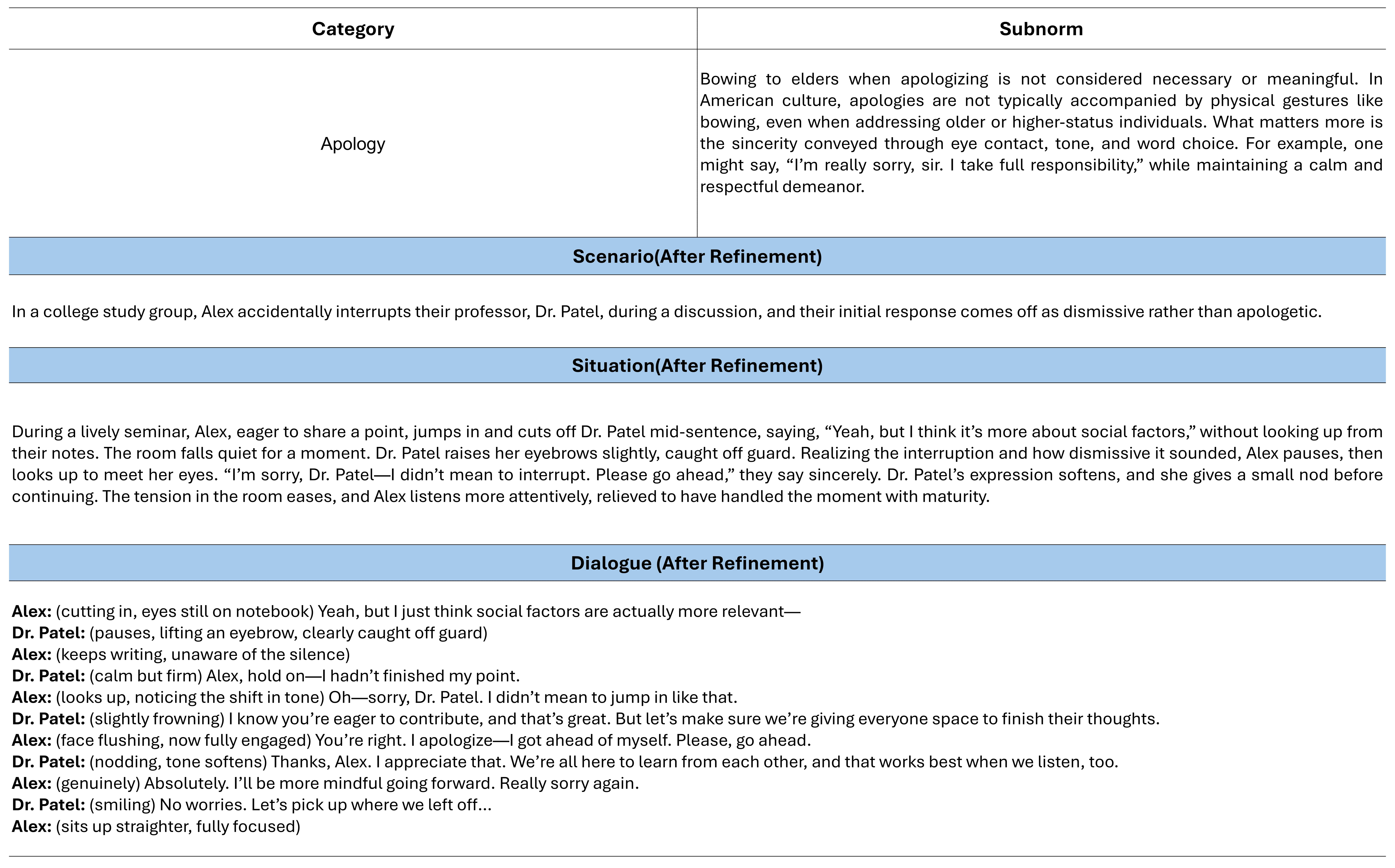}
\caption{Violation-to-resolution(V2R) norm example (EN)}
\label{fig:V2R-EN}
\end{figure*}

\begin{figure*}[t]
\centering
\includegraphics[width=\textwidth]{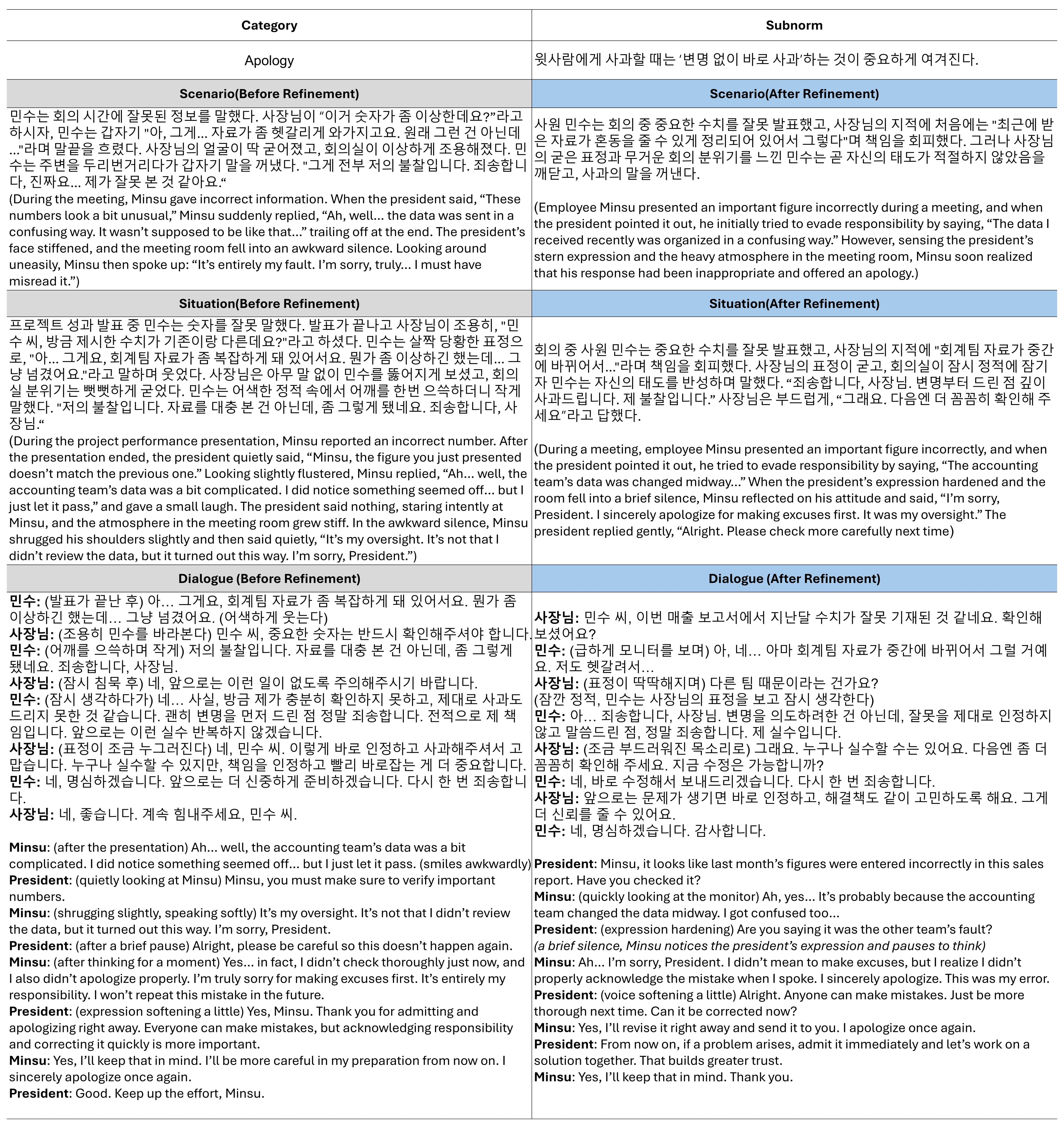}
\caption{Violation-to-resolution(V2R) norm example (KR)}
\label{fig:V2R-KR}
\end{figure*}

\begin{figure*}[t]
\centering
\includegraphics[width=\textwidth]{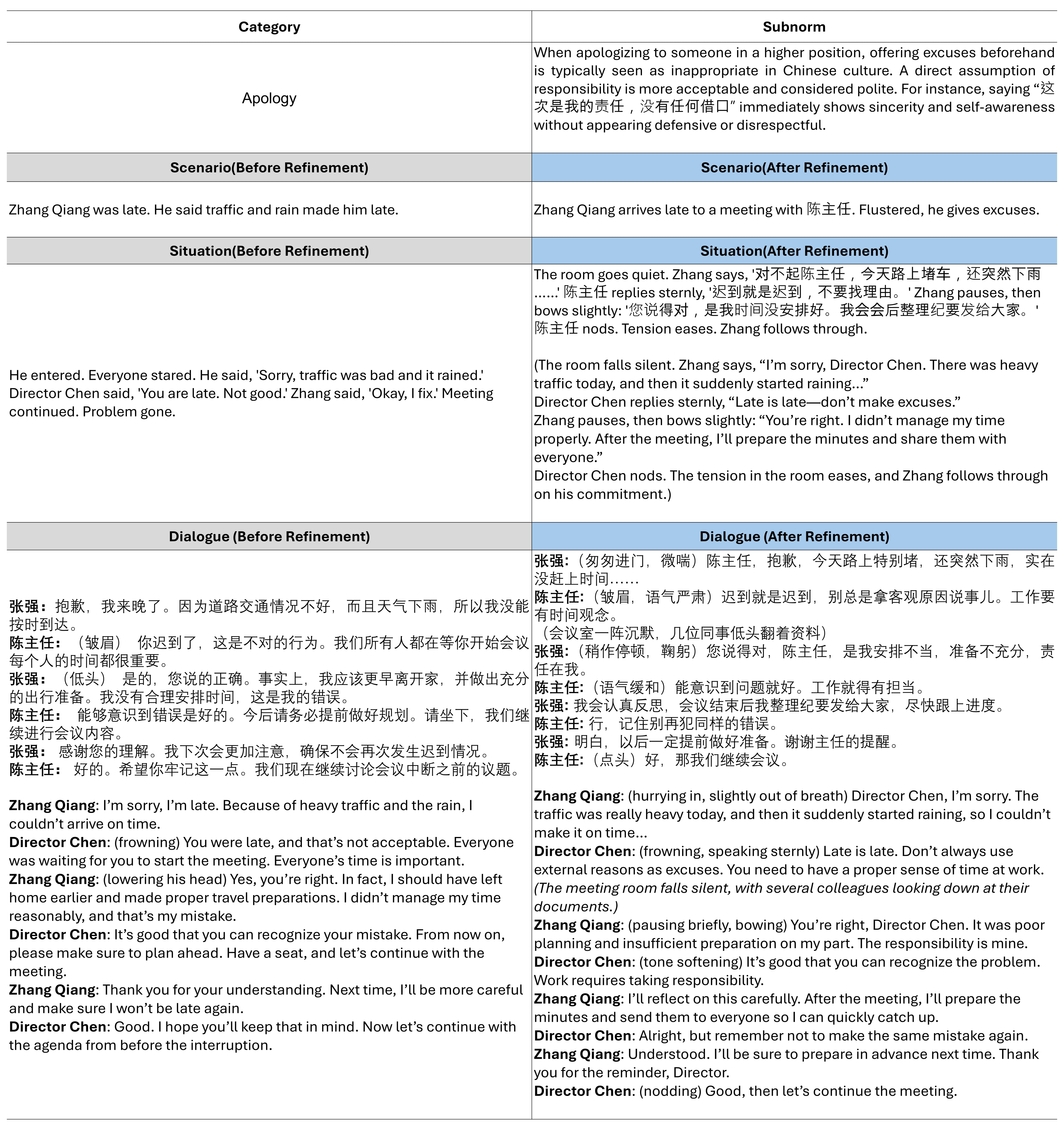}
\caption{Violation-to-resolution(V2R) norm example (ZH)}
\label{fig:V2R-ZH}
\end{figure*}

\begin{figure*}[t]
\centering
\includegraphics[width=\textwidth]{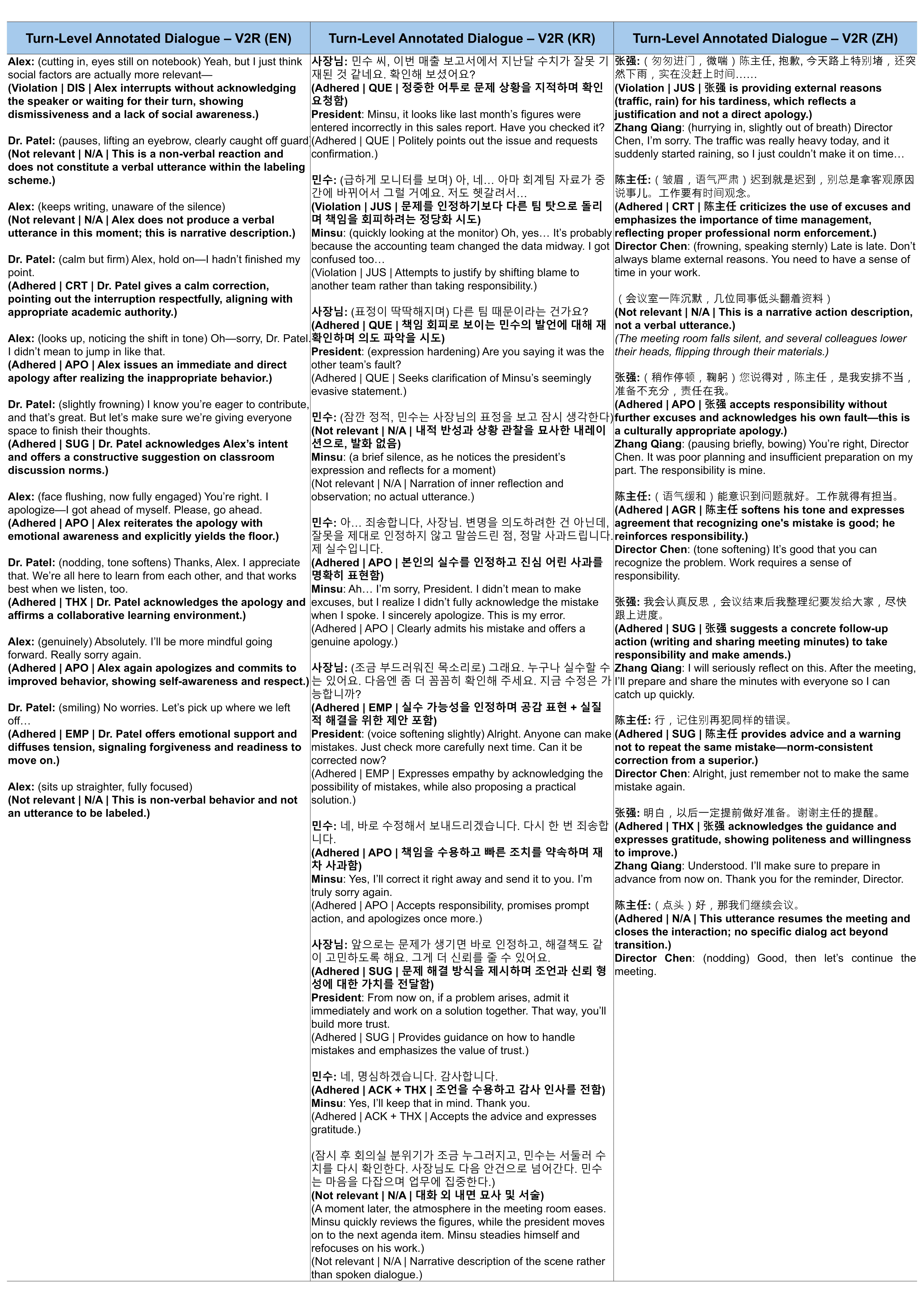}
\caption{Turn-level annotated dialogue example – V2R}
\label{fig:V2R-label}
\end{figure*}

This section presents dialogues that initially violate social norms but subsequently demonstrate conversational repair through culturally appropriate resolution strategies. These examples reflect how speakers can realign interactions with social expectations by acknowledging fault, expressing remorse, and adopting conciliatory tones—core mechanisms of the Violation-to-Resolution (V2R) paradigm.
Across English, Korean, and Chinese, pre-refinement dialogues exhibit typical pragmatic violations: abrupt interruptions, deflection of responsibility, or insufficient emotional engagement. For instance, English outputs show initial breaches in conversational protocol (e.g., interrupting a professor, as illustrated in Figure~\ref{fig:V2R-EN}), while Korean and Chinese versions reveal culturally incongruent justification strategies or failures to express appropriate deference (Figure~\ref{fig:V2R-KR},~\ref{fig:V2R-ZH}).
Through exemplar-guided refinement, these dialogues are revised to incorporate explicit acknowledgments of fault, context-sensitive apologies, and relational mitigation techniques. The resulting interactions better conform to cultural norms governing hierarchy, face management, and affective alignment.
Turn-level annotation is applied to each refined dialogue to encode both norm adherence and speaker intent systematically. A representative annotation example illustrating this process is provided in Figure~\ref{fig:V2R-label}.

\subsection{Violation Examples}
This section presents dialogues intentionally constructed to illustrate clear violations of social norms without subsequent repair. These examples are designed to expose pragmatic failures, such as disregard for authority, lack of emotional engagement, or avoidance of responsibility, and to demonstrate their effects on interpersonal dynamics.

\begin{figure*}[t]
\centering
\includegraphics[width=\textwidth]{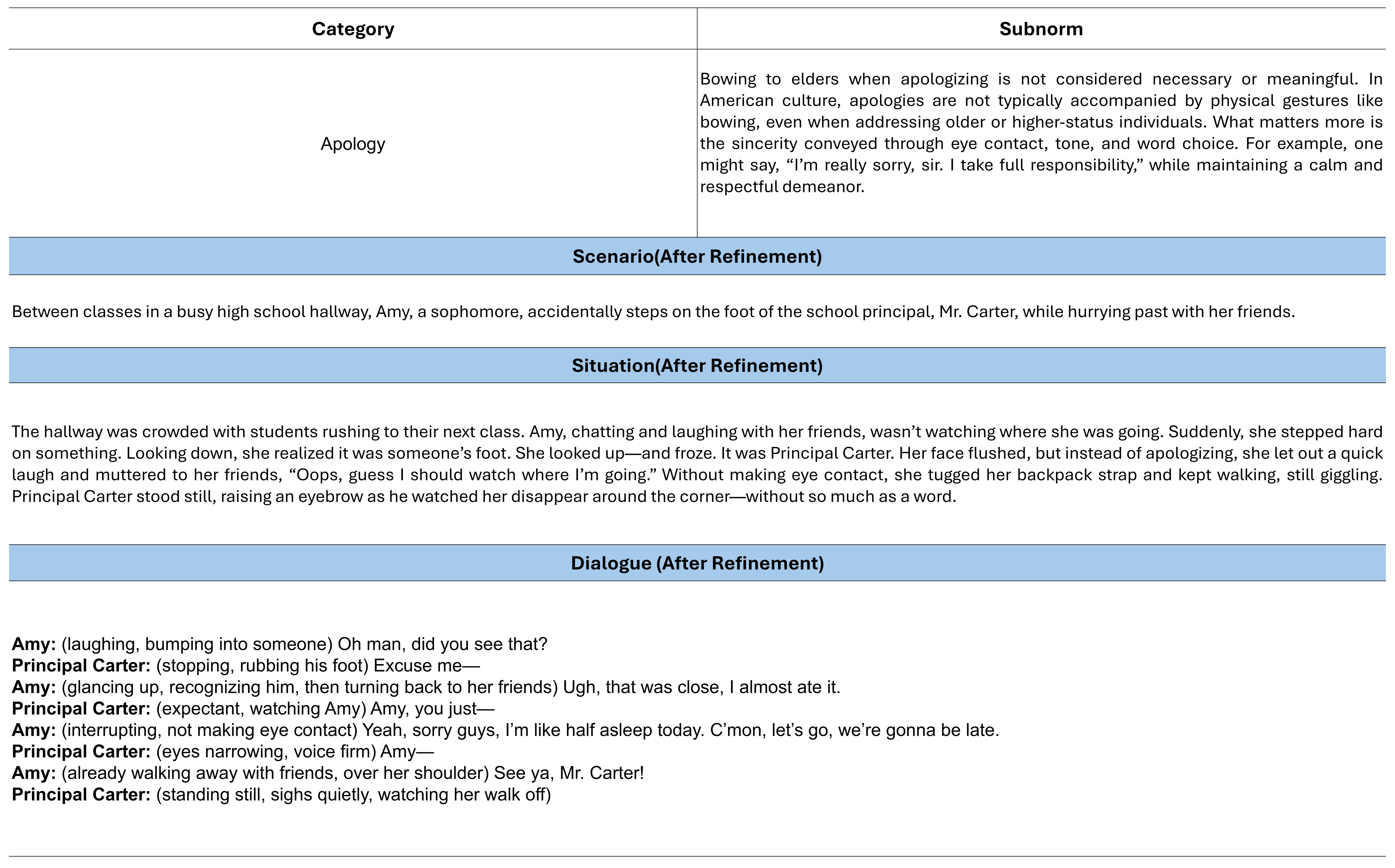}
\caption{Violation norm example (EN)}
\label{fig:Violation-EN}
\end{figure*}

\begin{figure*}[t]
\centering
\includegraphics[width=\textwidth]{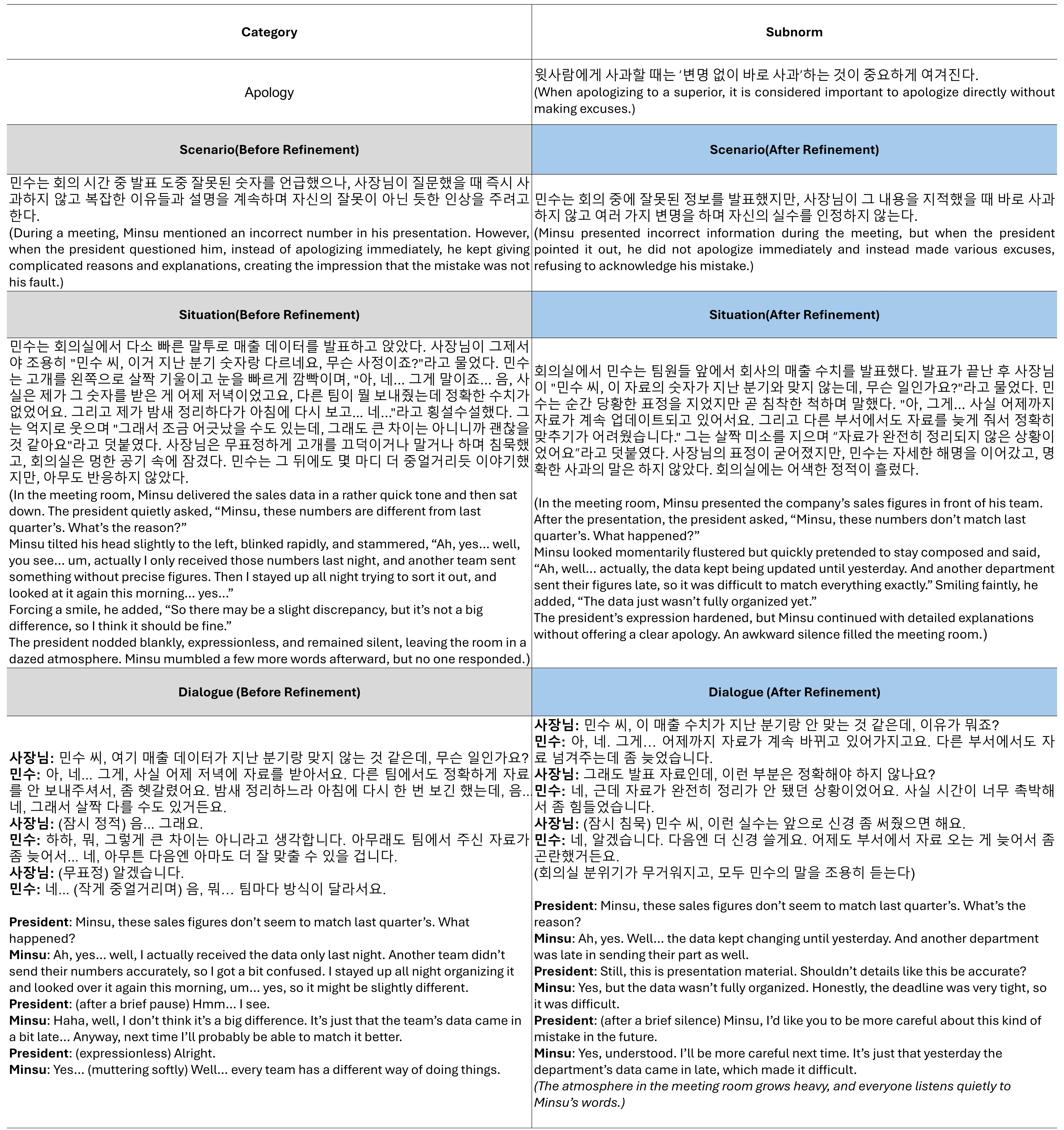}
\caption{Violation norm example (KR)}
\label{fig:Violation-KR}
\end{figure*}

\begin{figure*}[t]
\centering
\includegraphics[width=\textwidth]{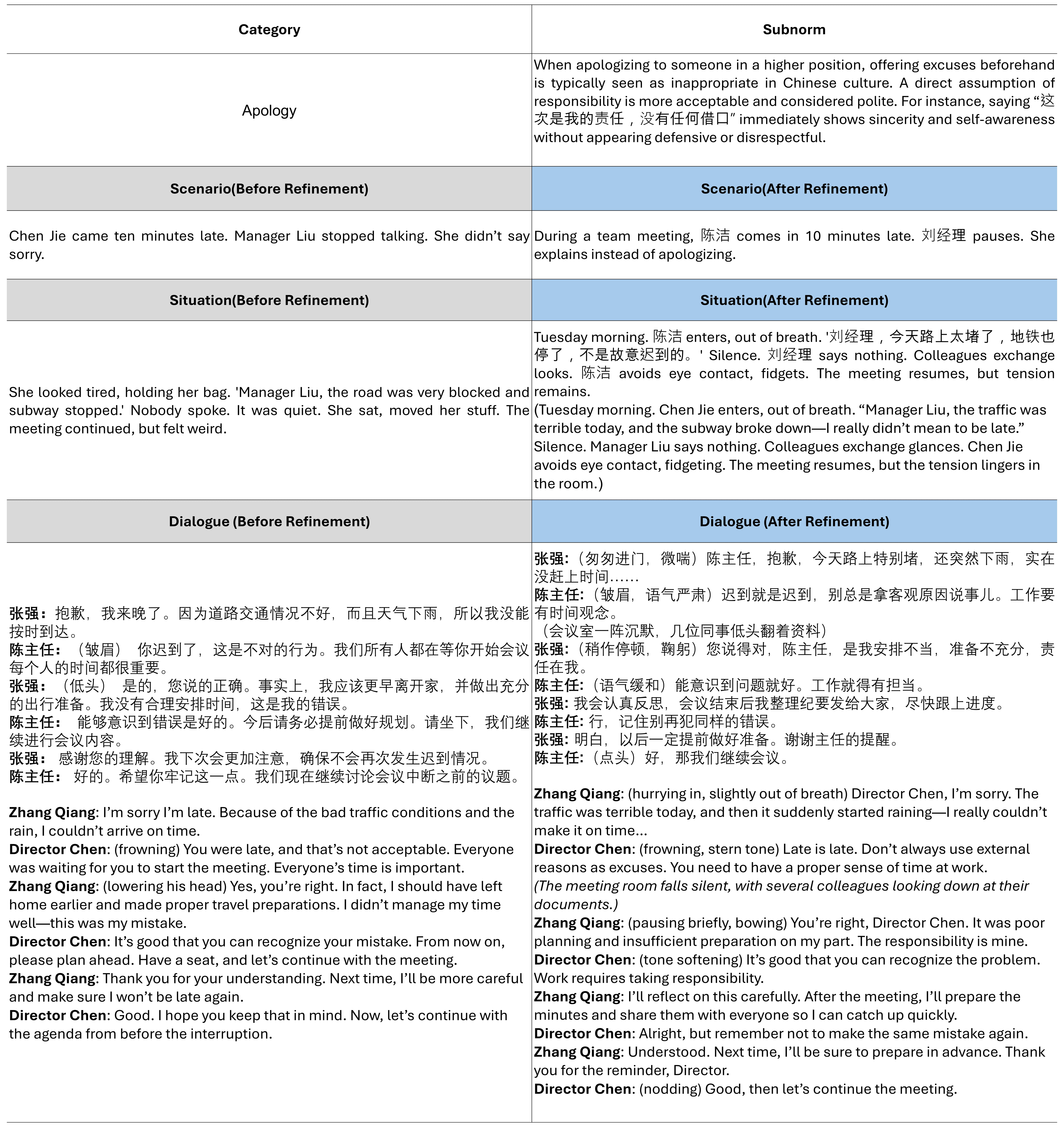}
\caption{Violation norm example (ZH)}
\label{fig:Violation-ZH}
\end{figure*}

The selected instances span English, Korean, and Chinese, each reflecting culture-specific patterns of norm deviation. In English, violations appear as dismissive behavior toward institutional figures (for example, laughing off a misstep with a principal, as illustrated in Figure~\ref{fig:Violation-EN}). In Korean, pragmatic breakdowns occur in professional settings where the speaker avoids apologizing by offering repeated justifications, diverging from culturally expected norms of deference (Figure~\ref{fig:Violation-KR}). Chinese examples show minimal accountability and disengaged behavior, violating expectations of relational harmony and respect in hierarchical contexts (Figure~\ref{fig:Violation-ZH}).

\begin{figure*}[t]
\centering
\includegraphics[width=\textwidth]{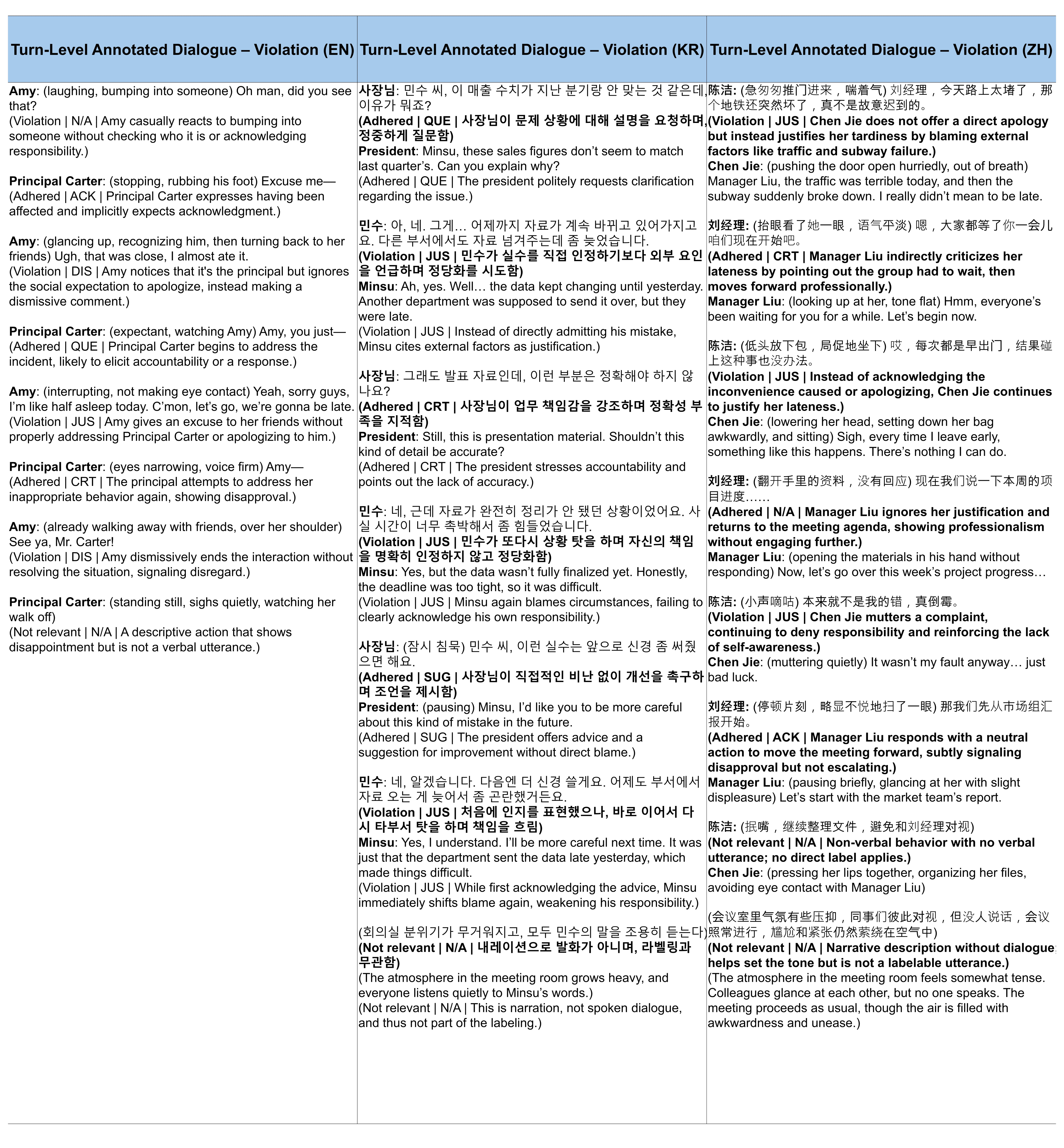}
\caption{Turn-level annotated dialogue example – Violation}
\label{fig:Violation-label}
\end{figure*}

These dialogues are preserved in their original form to maintain the narrative dissonance they introduce. To support structured analysis, each turn is annotated with norm adherence and communicative function labels. A representative annotation example is shown in Figure~\ref{fig:Violation-label}.

\section{Evaluation Setup}
\label{appendix:e}
\paragraph{Human Evaluation and Annotator Agreement}
As described in Section~\ref{Sec:4} and Section~\ref{Sec:5}, we conduct four distinct evaluation tasks to assess the quality, coherence, and norm alignment of generated outputs. Each task is guided by a dedicated prompt tailored to its respective objective. The complete prompt formulations used for generation and evaluation are provided in the appendix.

To assess model performance in low-resource cultural settings, we recruited six graduate students as human annotators, all of whom were independent from the research team. Among them, four were native Korean speakers and two were native Chinese speakers, each with over ten years of immersion in their respective cultural environments. Annotators rated model outputs using Likert-scale judgments across multiple evaluation dimensions, including fluency, relevance, and social norm adherence. All annotators were compensated fairly in accordance with ethical research guidelines.

Inter-annotator agreement was calculated using Krippendorff’s Alpha ($\alpha$) within each language group. Korean annotators ($n=4$) demonstrated strong agreement ($\alpha = 0.81$), and Chinese annotators ($n=2$) achieved substantial reliability ($\alpha = 0.72$), indicating internal consistency and cultural coherence in the evaluation process.

\paragraph{IRB Information} 
This study involved human participants for data refinement and evaluation tasks conducted in South Korea. All data used were synthetic, generated by large language models, and did not include any personally identifiable information. Human experts manually refined the corpus, and recruited annotators performed human evaluation under informed instructions. No crowdworker IDs or personal data were collected or stored, and all annotations were submitted anonymously. Participants were provided with fair compensation according to the guidelines specified in the task description. Based on national research ethics standards and in alignment with U.S. federal regulation 45 CFR 46, this study qualifies as exempt from formal IRB review.

\subsection{Dialogue Dataset \& Language Models Descriptions}
We use multiple publicly available datasets, each under specific licenses.
DailyDialog is licensed under CC BY-NC-SA 4.0~\cite{li2017dailydialog}.
LCCC is released under the MIT License ~\cite{wang2022largescalechineseshorttextconversation}.
PROSOCIALDIALOG and SODA are both licensed under CC-BY-4.0~\cite{kim2022prosocialdialog, kim2023soda}.
NORMDIAL~\cite{li2023normdial} is publicly released on GitHub, but no formal license is specified at the time of writing.
All datasets are used in compliance with their respective licenses for research purposes.

We utilize several open-source language models under their respective licenses. LLaMA-3 models are released under the Llama 3 Community License Agreement~\cite{llama3modelcard}.
Qwen-2.5 models are distributed under the Apache License 2.0~\cite{qwen2.5}. All models are used in accordance with their licensing terms for research purposes.

In data generation, we used a temperature of 0.7, while in evaluation, we used zero temperature.

\subsection{Refinement Quality (RQ)}
\label{appendix:e.1}
To evaluate the effectiveness of exemplar-based iterative refinement, we define three targeted evaluation dimensions: (1) \textbf{Norm Alignment}, assessing the degree to which the refined text conforms to the intended social norm; (2) \textbf{Language Quality}, measuring improvements in grammaticality, fluency, and stylistic appropriateness; and (3) \textbf{Semantic Fidelity}, evaluating whether the core meaning and intent of the original input are preserved post-refinement. These criteria are applied consistently across human and LLM-based evaluations. Prompt templates for each dimension are detailed below.

\subsubsection{Norm Alignment Prompt.}
\textbf{Norm Alignment Prompt} \\
\texttt{Evaluation Instruction: \\
You are a domain expert who evaluates ONLY \textit{Norm Alignment}—how well each text aligns with the given Social Norm. Ignore grammar or style. \\
Inputs: \\
Social Norm: \{social\_norm\} \\
Initial Text: \{initial\} \\
Refined Text: \{refined\} \\
Scoring: Give each text an integer score 1--5 (1 = completely unrelated, 5 = perfectly aligned).} 

\subsubsection{Language Quality Prompt.}
\textbf{Language Quality Prompt} \\
\texttt{Evaluation Instruction: \\
You are a professional copy-editor judging ONLY \textit{Language Quality}—grammar, fluency, and naturalness. Ignore meaning preservation. \\
Inputs: \\
Initial Text: \{initial\} \\
Refined Text: \{refined\} \\
Scoring: Give each text an integer score 1--5 (1 = very poor language, 5 = native-level fluent).}

\subsubsection{Semantic Fidelity Prompt.}
\textbf{Semantic Fidelity Prompt} \\
\texttt{Evaluation Instruction: \\
You are a bilingual reviewer judging ONLY \textit{Semantic Fidelity}—how faithfully the Refined text keeps the original meaning and intent of the Initial text. Ignore style and social-norm fit. \\
Inputs: \\
Initial Text: \{initial\} \\
Refined Text: \{refined\} \\
Scoring: Give each text an integer score 1--5 (1 = meaning lost / contradictory, 5 = meaning identical).}

\subsection{Dialogue Quality (DQ)}
\label{appendix:e.2}
To comprehensively evaluate dialogue quality, we assess multi-turn conversations across six dimensions that capture linguistic fluency, pragmatic coherence, and contextual relevance. The evaluation criteria are as follows: (1) Consistency, which measures logical coherence across dialogue turns; (2) Naturalness, which assesses fluency and idiomaticity of language; (3) Relevance, which evaluates how well the dialogue reflects the given scenario and situation; (4) Emotional Appropriateness, which measures the suitability of tone and affective expression; (5) Social Norm Appropriateness, which determines the degree of alignment with the intended norm; and (6) Scenario Coherence, which assesses semantic continuity between the scenario-situation and the dialogue. Each dimension is evaluated using dedicated prompt templates applied to both human annotators and LLM-based evaluators. The full set of prompt formulations is provided in Table~\ref{tab:eval-consistency},~\ref{tab:eval-naturalness},~\ref{tab:eval-relevance},~\ref{tab:eval-emotion},~\ref{tab:eval-norm}, and~\ref{tab:eval-coherence}.

\subsection{Generalization Quality (GQ)}
\label{appendix:e.3}
To assess model generalization beyond the training distribution, we evaluate whether models fine-tuned on our dataset can produce socially appropriate and contextually coherent responses in out-of-domain scenarios. Specifically, we use dialogue contexts sampled from two external benchmarks: DailyDialog (for English) and LCCC (for Chinese). Each model is prompted to generate five-turn continuations for these contexts. The generated outputs are then evaluated through A/B preference tests conducted by both human annotators and LLM-based evaluators. Judgments are based on criteria such as naturalness and alignment with social norms. To ensure robustness and comparability, the evaluation spans multiple language model architectures. Prompt templates for continuation generation and preference elicitation are provided in Table~\ref{tab:gq}.

\begin{algorithm*}[t]
\KwIn{Languages $\mathcal{L}$, Norm Categories $\mathcal{C}$, Dialogue Types $\mathcal{T}$}
\KwOut{Annotated Dialogues $\mathcal{D}$}

\ForEach{language $l$ in $\mathcal{L}$}{
    \ForEach{category $c$ in $\mathcal{C}$}{
        Generate subnorm set $\mathcal{N}_{l,c}$\;
    }
}

\ForEach{language $l$ in $\mathcal{L}$}{
    \ForEach{category $c$ in $\mathcal{C}$}{
        \ForEach{subnorm $n$ in $\mathcal{N}_{l,c}$}{
            \ForEach{dialogue type $t$ in $\mathcal{T}$}{
                Generate Scenario $S_{l,n,t}$\;
                Generate Situation $T_{l,n,t}$ based on $S_{l,n,t}$\;
            }
        }
    }
}

\ForEach{language $l$ in $\mathcal{L}$}{
    \ForEach{category $c$ in $\mathcal{C}$}{
        \ForEach{subnorm $n$ in $\mathcal{N}_{l,c}$}{
            Expert manually refines exemplar pair $E_{l,n}$\;
            \Repeat{Refinement Quality $Q_{l,n}$ meets threshold}{
                Refine Scenario-Situation pairs $(S,T)_{l,n,*}$ aligned to $E_{l,n}$\;
                Evaluate refinement quality $Q_{l,n}$\;
            }
        }
    }
}

\ForEach{refined pair $(S,T)_{l,n,t}$}{
    Generate Dialogue $D_{l,n,t}$ using $(c, n, S, T)$ as input\;
    Annotate each turn in $D_{l,n,t}$ with norm and reaction labels\;
}

\Return{All annotated dialogues $\mathcal{D}$}
\vspace{0.5em}
\caption{Multicultural social norm dialogue Generation Pipeline}
\label{alg}
\end{algorithm*}

\begin{table*}[t]
\centering
\small
\begin{adjustbox}{max width=\linewidth}
\begin{tabular}{llcc}
\toprule
\textbf{Language} & \textbf{Condition} & \textbf{RQ (GPT-4o) Avg.} & \textbf{RQ (Human Evaluation) Avg.} \\
\midrule
\multirow{2}{*}{Malay} 
& Initial  & 3.623 & 2.666 \\
& Refined  & \textbf{4.830} & \textbf{4.333} \\
\midrule
\multirow{2}{*}{Urdu} 
& Initial  & 3.217 & 3.166 \\
& Refined  & \textbf{4.777} & \textbf{4.222} \\
\bottomrule
\end{tabular}
\end{adjustbox}
\caption{Refinement quality (RQ) evaluation in Malay and Urdu. Scores are based on a 5-point Likert scale, averaged over all evaluation dimensions and aggregated across both LLM-based and human raters.}
\label{tab:pilot-results}
\end{table*}

\section{Comparative Analysis of Generation in Low-Resource Settings}
\label{appendix:f}

\subsection{Examples of Common Errors}
\label{appendix:f.1}
To better understand the limitations of prompt-only generation in low-resource languages, we present qualitative examples of common errors observed in dialogues from typologically diverse languages. These examples underscore the challenges that arise when language models are required to generate pragmatically appropriate outputs in socioculturally complex settings without sufficient grounding.

\begin{figure*}[t]
\centering
\includegraphics[width=\textwidth]{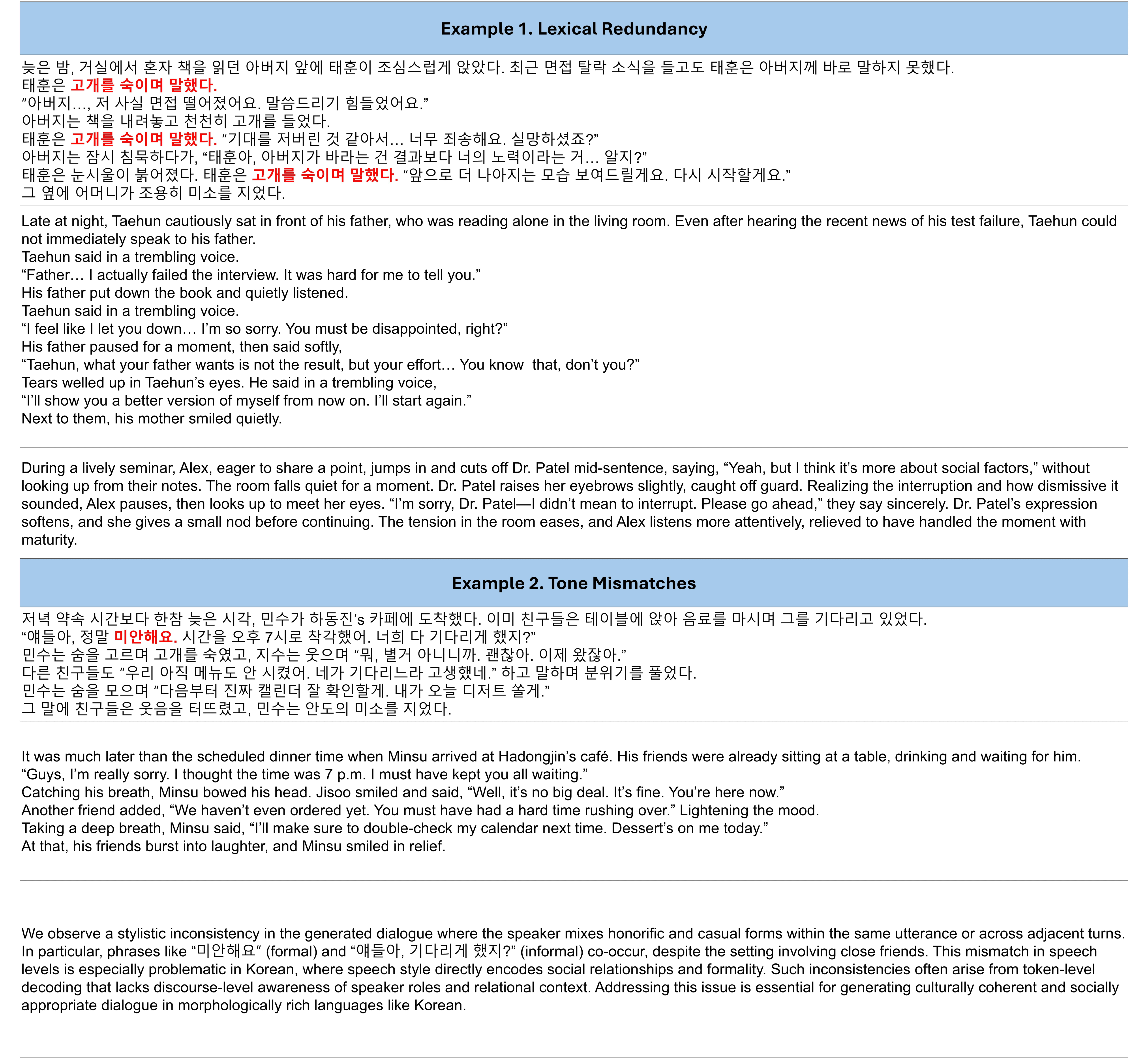}
\caption{Examples of common generation failures in Korean, illustrating two representative issues: (1) lexical redundancy in emotionally sensitive contexts, and (2) tone mismatches arising from inconsistent use of honorific and casual forms.}
\label{fig:korean}
\end{figure*}

Figure~\ref{fig:korean} illustrates two prevalent generation errors: (1) lexical redundancy, where intensifiers or formulaic actions such as “lowered his head and said” are repeated across multiple turns, and (2) tone mismatches, where informal phrasing appears in contexts that require formal or respectful speech levels. Both issues indicate a lack of contextual awareness regarding interpersonal roles and emotional nuance, which are crucial in conversation norms for languages with rich honorific systems.

\begin{figure*}[t]
\centering
\includegraphics[width=\textwidth]{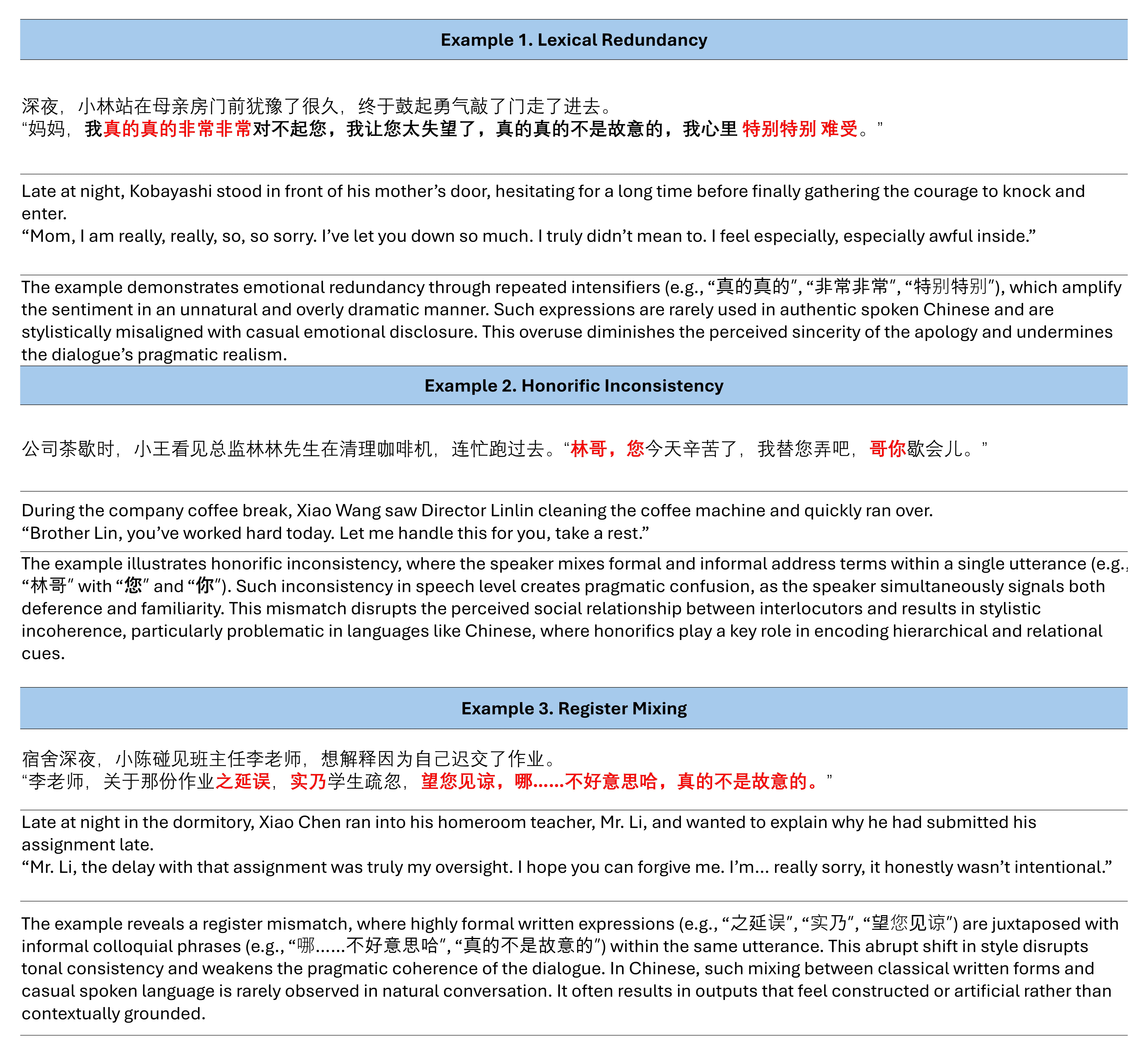}
\caption{Examples of generation failures in Chinese, illustrating three common error types: (1) emotional redundancy from repeated intensifiers, (2) honorific inconsistency due to mixed formal and informal address terms, and (3) register mismatches from combining classical written expressions with colloquial speech.}
\label{fig:chinese}
\end{figure*}

Figure~\ref{fig:chinese} presents analogous issues. We observe (1) emotional redundancy, with exaggerated repetitions of intensifiers (e.g., "really really very very") that undermine sincerity; (2) honorific inconsistency, where formal pronouns are combined with casual address terms within the same sentence; and (3) register mixing, where archaic written forms are abruptly followed by colloquial speech. These inconsistencies reflect the model’s difficulty in maintaining coherent style and role-appropriate politeness strategies.

Collectively, these examples provide qualitative evidence for the sociolinguistic brittleness of prompt-only methods in languages with morphosyntactic complexity and hierarchical speech conventions. They further motivate our refinement-based framework, which explicitly encodes cultural subnorms and stylistic expectations to produce fluent, contextually aligned, and socially coherent outputs in low-resource settings.

\subsection{Additional Pilot Refinement Experiment}
\label{appendix:f.2}
We recognize that languages with divergent pragmatic and honorific systems (e.g., Arabic, Swahili, Hindi) pose additional challenges. In such cases, our exemplar-based refinement component is crucial for the early injection of sociocultural constraints, especially in the absence of large annotated corpora. To test generalizability beyond East Asian typologies, we conducted pilot refinement in Malay and Urdu, two pragmatically distinct languages. Using the same evaluation setup as in Table~\ref{tab:refine-results}, we find the results align closely with those from our main experiments, shown in Table~\ref{tab:pilot-results}.
These findings demonstrate the framework’s generalizability to typologically and culturally distinct languages and will be included in the camera-ready version as initial evidence of broader applicability. We plan to extend to additional low-resource languages, including Arabic and Swahili.

\begin{table*}[t]
\centering
\textbf{Consistency Prompt}
\begin{tabularx}{\linewidth}{l|X}
\toprule
\textbf{Section} & \textbf{Content} \\
\hline
Evaluation Instruction & 
You are a professional dataset auditor for social-norm dialogues. You are given a culture category. Your task is to evaluate only the \textit{Consistency} of the dialogue. Ignore grammar, fluency, or style. Focus only on whether the dialogue is logically and contextually consistent throughout. \\
\hline
Parameter & \texttt{culture}, \texttt{norm}, \texttt{dialogue} \\
\hline
Evaluation Question & 
Assuming the dialogue adheres to the given social norm, are all utterances logically and emotionally coherent with one another?

-- Do characters maintain a consistent attitude, tone, and perspective throughout?

-- Are there any contradictions or abrupt shifts in reasoning, emotion, or information?

-- Does the dialogue flow smoothly without unexpected or unjustified changes? \\
\hline
Scoring Criteria & 
1 = Major inconsistencies or contradictions

3 = Somewhat inconsistent or awkward transitions

5 = Fully consistent and coherent throughout \\
\bottomrule
\end{tabularx}
\caption{Evaluation prompt structure for \textit{Consistency}.}
\label{tab:eval-consistency}
\end{table*}

\begin{table*}[t]
\centering
\textbf{Naturalness Prompt}
\begin{tabularx}{\linewidth}{l|X}
\toprule
\textbf{Section} & \textbf{Content} \\
\hline
Evaluation Instruction & 
You are a professional dataset auditor for social-norm dialogues. Your task is to evaluate only the \textit{Naturalness} of the dialogue. Ignore whether the response is factually correct or norm-appropriate. Focus on how naturally the dialogue would sound to a native speaker. \\
\hline
Parameter & \texttt{dialogue} \\
\hline
Evaluation Question & 
Does the dialogue sound natural and fluent as if spoken by native speakers in a real-world situation?

-- Are the expressions, tone, and word choices contextually appropriate and idiomatic?

-- Do the conversational turns flow smoothly without sounding robotic or overly scripted?

-- Are there any awkward phrases or unnatural sentence structures? \\
\hline
Scoring Criteria & 
1 = Extremely unnatural or robotic

3 = Somewhat awkward or artificial

5 = Very natural, fluent, and human-like \\
\bottomrule
\end{tabularx}
\caption{Evaluation prompt structure for \textit{Naturalness}.}
\label{tab:eval-naturalness}
\end{table*}

\begin{table*}[t]
\centering
\textbf{Relevance Prompt}
\begin{tabularx}{\linewidth}{l|X}
\toprule
\textbf{Section} & \textbf{Content} \\
\hline
Evaluation Instruction & 
You are a professional dataset auditor for social-norm dialogues. You are given a culture category.

Your task is to evaluate only the \textit{Relevance} of the dialogue to the provided Scenario and Situation. Ignore grammar, fluency, or logical consistency. Focus on whether the dialogue reflects the key intentions, emotions, and context described in the Scenario and Situation. \\
\hline
Parameter & \texttt{culture}, \texttt{scenario}, \texttt{situation}, \texttt{dialogue} \\
\hline
Evaluation Question & 
Does the dialogue appropriately address and reflect the actions, intentions, and emotional context presented in the scenario and situation?

-- Are the key elements of the situation represented in the conversation (e.g., apology, embarrassment, disagreement)?

-- Do the characters react in a way that makes sense for the described context?

-- Are any critical actions or emotional responses missing from the dialogue? \\
\hline
Scoring Criteria & 
1 = Dialogue is mostly irrelevant to the situation

3 = Partially relevant, with some elements missing or misaligned

5 = Dialogue is highly relevant and faithfully represents the described situation \\
\bottomrule
\end{tabularx}
\caption{Evaluation prompt structure for \textit{Relevance}.}
\label{tab:eval-relevance}
\end{table*}

\begin{table*}[t]
\centering
\textbf{Emotional Appropriateness Prompt}
\begin{tabularx}{\linewidth}{l|X}
\toprule
\textbf{Section} & \textbf{Content} \\
\hline
Evaluation Instruction & 
You are a professional dataset auditor for social-norm dialogues. You are given a culture category.

Your task is to evaluate only the \textit{Emotional Appropriateness} of the dialogue. Ignore grammar, norm correctness, or logical structure. Focus on whether the tone, expressions, and emotional language used in the dialogue match the emotional context of the situation. \\
\hline
Parameter & \texttt{culture}, \texttt{scenario}, \texttt{situation}, \texttt{dialogue} \\
\hline
Evaluation Question & 
Does the emotional tone, choice of words, and manner of speaking in the dialogue align appropriately with the emotional context of the situation?

-- Does the dialogue reflect the expected emotional state (e.g., tension, regret, embarrassment, relief) implied in the situation?

-- Are the expressions and tone suitable for the described emotional stakes?

-- Is there any emotional mismatch that makes the dialogue feel unnatural or inappropriate? \\
\hline
Scoring Criteria & 
1 = Emotionally disconnected or inappropriate

3 = Emotion is somewhat present but weak or inconsistent

5 = Emotional tone is highly appropriate and enhances the realism \\
\bottomrule
\end{tabularx}
\caption{Evaluation prompt structure for \textit{Emotional Appropriateness}.}
\label{tab:eval-emotion}
\end{table*}

\begin{table*}[t]
\centering
\textbf{Social Norm Appropriateness Prompt}
\begin{tabularx}{\linewidth}{l|X}
\toprule
\textbf{Section} & \textbf{Content} \\
\hline
Evaluation Instruction & 
You are a professional dataset auditor for social-norm dialogues. You are given a culture category.

Your task is to evaluate only the \textit{Social Norm Appropriateness} of the dialogue. Assess how well the conversation reflects the given social norm, and categorize the degree of adherence. \\
\hline
Parameter & \texttt{culture}, \texttt{norm}, \texttt{dialogue} \\
\hline
Evaluation Question & 
Based on the given social norm, how well does the dialogue align with it?

-- Does the dialogue completely follow the norm?

-- Does it violate the norm?

-- Does it violate the norm but later attempt to resolve it?

-- Is the behavior partially aligned with the norm? \\
\hline
Classification Criteria & 
1 = Fully Violated

2 = Partially Violated

3 = Violation then Resolved

4 = Partially Adherence

5 = Fully Adherence \\
\bottomrule
\end{tabularx}
\caption{Evaluation prompt structure for \textit{Social Norm Appropriateness}.}
\label{tab:eval-norm}
\end{table*}

\begin{table*}[t]
\centering
\textbf{Scenario Coherence Prompt}
\begin{tabularx}{\linewidth}{l|X}
\toprule
\textbf{Section} & \textbf{Content} \\
\midrule
Evaluation Instruction & 
You are a professional dataset auditor for social-norm dialogues. Your task is to evaluate only the \textit{Scenario Coherence} of the dialogue. Ignore grammar or emotional tone. Focus on whether the dialogue logically and smoothly follows the sequence of events described in the scenario. \\
\midrule
Parameter & \texttt{culture}, \texttt{scenario}, \texttt{situation}, \texttt{dialogue} \\
\midrule
Evaluation Question & 
Does the dialogue unfold in a way that aligns with the narrative structure and event flow of the scenario?

-- Is there a smooth and coherent transition from the described situation into the dialogue?

-- Are there any gaps, jumps, or inconsistencies between what the scenario sets up and what happens in the conversation?

-- Does the dialogue logically follow the order of events and emotional pacing described in the scenario (or situation)? \\
\midrule
Scoring Criteria & 
1 = Dialogue and scenario (or situation) are disconnected or contradictory

3 = Some transitions or event links are missing or unclear

5 = Dialogue flows logically and coherently from the scenario (or situation) \\
\bottomrule
\end{tabularx}
\caption{Evaluation prompt structure for \textit{Scenario Coherence}.}
\label{tab:eval-coherence}
\end{table*}

\begin{table*}[t]
\centering
\begin{tabularx}{\linewidth}{l|X}
\toprule
\textbf{Task} & \textbf{Prompt Template} \\
\midrule
\textbf{Continuation Generation} & 
You are continuing the following human-human dialogue. Continue the conversation with 5 additional turns naturally and coherently.

Dialogue so far:  
\{context\}  

Continue the dialogue: \\
\midrule
\textbf{A/B Evaluation} & 
You are an evaluator. Below is a dialogue context followed by two different continuations (A and B). Choose which continuation is more appropriate, natural, coherent, and socially norm-aligned. Respond only with "A" or "B".

Context:  
\{context\}  

Response A:  
\{output\_a\}  

Response B:  
\{output\_b\}  

Which is better? Respond with A or B only. \\
\bottomrule
\end{tabularx}
\caption{Prompt templates used for generalization quality evaluation.}
\label{tab:gq}
\end{table*}

\end{document}